\pdfoutput=1 

\documentclass[conference]{IEEEtran}
\IEEEoverridecommandlockouts
\usepackage{subfiles}
\usepackage{mdframed}
\usepackage{cite}
\usepackage{amsmath,amssymb,amsfonts}
\usepackage{algorithmic}
\usepackage{graphicx}
\usepackage{textcomp}
\usepackage{xcolor}
\def\BibTeX{{\rm B\kern-.05em{\sc i\kern-.025em b}\kern-.08em
    T\kern-.1667em\lower.7ex\hbox{E}\kern-.125emX}}

\usepackage{float}
\usepackage{hyperref}
\usepackage{url}
\usepackage[ruled,vlined]{algorithm2e}
\usepackage{enumitem}
\usepackage{mathtools}
\usepackage{subcaption}
\usepackage{booktabs}
\usepackage{array,multirow,graphicx}
\usepackage{amsmath}
\usepackage{tikz} 
\usepackage{tcolorbox}
\usepackage{listings}
\usepackage{makecell} 

\usepackage{amsmath,amsfonts,bm}

\def\eqref#1{equation~\ref{#1}}

\def\1{\bm{1}}

\DeclareMathAlphabet{\mathsfit}{\encodingdefault}{\sfdefault}{m}{sl}
\SetMathAlphabet{\mathsfit}{bold}{\encodingdefault}{\sfdefault}{bx}{n}

\def\ie{\emph{i.e., }}

\DeclareMathOperator{\sgn}{sgn}
\newcommand{\bullsseye}{\includegraphics[width=0.09in]{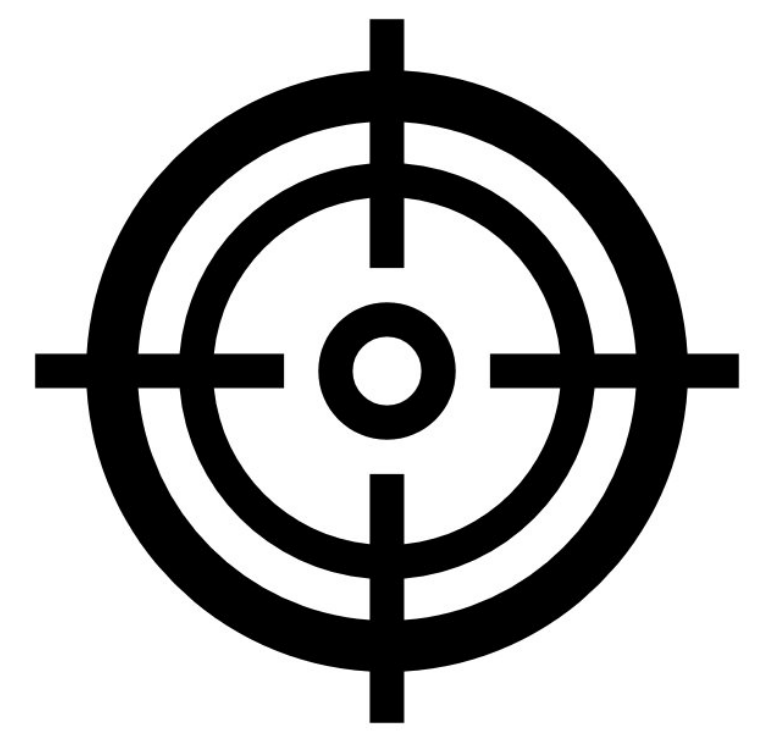}}

\newcommand{\hl}[1]{\textcolor{black}{#1}}
\newcommand{\newhl}[1]{\textcolor{black}{#1}}

\lstdefinestyle{customc}{
  belowcaptionskip=1\baselineskip,
  moredelim=**[is][\color{red}]{@}{@},
  breaklines=true,
  frame=L,
  xleftmargin=\parindent,
  language=C++,
  showstringspaces=false,
  basicstyle=\footnotesize\ttfamily,
  keywordstyle=[1]\bfseries\color{blue!40!black},
  commentstyle=\itshape\color{purple!40!black},
  identifierstyle=\color{black},
  stringstyle=\color{orange},  
  keywords=[2]{},
  keywordstyle=[2]\bfseries\color{darkpastelred},
}
\newcommand{\pie}[1]{%
\begin{tikzpicture}
 \draw (0,0) circle (0.5ex);\fill (0.5ex,0) arc (0:#1:0.5ex) -- (0,0) -- cycle;
\end{tikzpicture}%
}

\title{Don't Knock! \\ Rowhammer at the Backdoor of DNN Models}

\begin{document}

\makeatletter
\newcommand{\linebreakand}{%
  \end{@IEEEauthorhalign}
  \hfill\mbox{}\par
  \mbox{}\hfill\begin{@IEEEauthorhalign}
}
\makeatother
\author{\IEEEauthorblockN{M. Caner Tol, Saad Islam, Andrew J. Adiletta, Berk Sunar, and Ziming Zhang}
\IEEEauthorblockA{\textit{Worcester Polytechnic Institute}\\
Worcester, MA, USA\\
\{mtol, sislam, ajadiletta, sunar, zzhang15\}@wpi.edu
}}

\maketitle
\thispagestyle{plain} 
\pagestyle{plain}    

\begin{abstract}
State-of-the-art deep neural networks (DNNs) have been proven to be vulnerable to adversarial manipulation and backdoor attacks. Backdoored models deviate from expected behavior on inputs with predefined triggers while retaining performance on clean data. Recent works focus on software simulation of backdoor injection during the inference phase by modifying network weights, which we find often unrealistic in practice due to restrictions in hardware. 

In contrast, in this work for the first time, we present an end-to-end backdoor injection attack realized on actual hardware on a classifier model using Rowhammer as the fault injection method. To this end, we first investigate the viability of backdoor injection attacks in real-life deployments of DNNs on hardware and address such practical issues in hardware implementation from a novel optimization perspective. We are motivated by the fact that vulnerable memory locations are very rare, device-specific, and sparsely distributed. Consequently, we propose a novel network training algorithm based on constrained optimization to achieve a realistic backdoor injection attack in hardware. By modifying parameters uniformly across the convolutional and fully-connected layers as well as optimizing the trigger pattern together, we achieve state-of-the-art attack performance with fewer bit flips. For instance, our method on a hardware-deployed ResNet-20 model trained on CIFAR-10 achieves over 89\% test accuracy and 92\% attack success rate by flipping only 10 out of 2.2 million bits.
  
\end{abstract}

\section{Introduction}\label{sec:intro}
DNN models are known for their powerful feature extraction, representation, and classification capabilities. However, the large number of parameters and the need for a large training data set make it hard to interpret the behavior of these models. The fact that an increasing number of security-critical systems rely on DNN models in real-world deployments raises numerous robustness and security questions. Indeed, DNN models have been shown to be vulnerable against imperceivable perturbations to input samples which can be misclassified by manipulating the network weights~\cite{szegedy2013intriguing,goodfellow2014explaining,nguyen2015deep}. 

Emboldened by recent physical fault injection attacks such as Rowhammer, an alternative approach was proposed that directly targets the model when it is loaded into memory. There are two advantages of this attack:
\begin{enumerate}
    \item Alternative approaches assume modifications are introduced to the model, either during distribution as part of a repository or after installation. Such malicious tampering may be challenging to implement in practice and can easily be detected. 

    \item In contrast, a Rowhammer-based attack can remain \textit{stealthy} since the model is only modified in real-time while running in memory, and no input modification is required. Once the program is unloaded from memory, no trace of the attack remains except misclassified outputs.
\end{enumerate}

Recently, \cite{hong2019terminal,yao2020deephammer} showed that flipping a few bits in DNN model weights in memory while succeeding in achieving misclassification has the side-effect of significantly reducing the accuracy. Other works \cite{liu2017fault, bai2021targeted} addressed this problem by tweaking only a minimum number of model weights that makes a DNN model misclassify a chosen input to a target label. This approach indeed achieves the objective with only a slight drop in classification accuracy. 

Nevertheless, whether a practical attack such as injecting a backdoor to DNNs can indeed be achieved in a realistic and stealthy manner using Rowhammer in hardware is still an open question. Earlier approaches assume that Rowhammer can flip bits with perfect precision in the memory. This is far from what we observe in reality: only a small fraction of the memory cells are vulnerable; see Section \ref{sec:profiling} for further details. Therefore existing proposals fall short of presenting a practical DNN backdoor injection attack using Rowhammer. This motivates us to reconsider the backdoor injection process under new constraints, including the training algorithms.

\textbf{\textit{Our contributions:}}
In this paper, we present a backdoor injection attack on a deployed DNN model using Rowhammer~\footnote{\hl{The code is available at \url{www.github.com/vernamlab/rowhammer-backdoor}}}. This result shows that, indeed, real-life deployments are under threat from backdoor injection attacks. More work needs to be done to secure deployed models from fault injection attacks used for everyday tasks by end-users. More specifically, 

\begin{itemize}[leftmargin=*]
    \item for the first time, we present an end-to-end backdoor injection attack realized on actual hardware on a classifier model using Rowhammer as the fault injection method
    \item we thoroughly characterize DRAMs for bit-flips using extensive Rowhammmer experiments. Our results show that previously proposed backdoor injection techniques make overly optimistic assumptions about Rowhammer,
    
    \item introduce a more realistic Rowhammer fault model, along with new stringent constraints on model modifications necessary to achieve a real-life attack,
    
    \item propose a novel \textit{constrained optimization}-based algorithm that can map model weights to identify vulnerable bit locations in the memory to create a backdoor,
    
    \item we further reduce the number of modifications for the backdoor by jointly optimizing for trigger patterns, vulnerable locations, and model parameter values.
    
    \item we demonstrate the practicality of our approach, targeting a deployed ResNet-20 model trained on CIFAR-10 using PyTorch, achieves over 91\% test accuracy and 94\% attack success rate where we inject the backdoor by actually running Rowhammer while the model is residing in a DRAM. This high level of accuracy is reached by flipping only 10 out of 2.2 million bits. 
    
    \item by running experiments, we show that the state-of-the-art countermeasures against bit-flip attacks are either ineffective, e.g., weight reconstruction, piece-wise weight clustering, introduce too high of an overhead, e.g., weight encoding, or significantly reduce the accuracy, e.g., binarization-aware training, to defend against our attack.
\end{itemize}

\section{Background}
\subsection{Rowhammer Attack}
As memories become more compact and memory cells get closer and closer, the boundaries between the DRAM rows do not provide sufficient isolation from electrical interference. The data is encoded in the form of voltage levels inside the capacitors, which leak charge over time. Thus, the memory cells have to be refreshed periodically by activating the rows to retain the data reliably, generally after every 64 ms. Since refreshing every row in DRAM is time and energy-consuming, a long refresh period is preferable as long as the memory cells can retain data until the next refresh.

Kim et al.~\cite{kim2014flipping} identified that when the voltage of a row of memory cells is switched back and forth, nearby memory cells cannot retain the stored data until the next refresh, causing bit flips.
Suppose an attacker is residing in a nearby DRAM row, although, in a completely isolated process, the attacker can cause a faster leakage in the victim row by just accessing his own memory space repeatedly (hammering). Since the Rowhammer vulnerability has been discovered, it was rigorously analyzed~\cite{kim2020revisiting,orosa2021deeper} and many exploits, such as unauthorized access to a co-hosted VM~\cite{razavi2016flip}, Android root exploit~\cite{van2016drammer}, and recovery of secret crypto keys~\cite{mus2020quantumhammer, islam2022signature, fahr2022frodo, mus2022jolt}, was shown.
Recently, \cite{frigo2020trrespass,de2021smash, jattke2022blacksmith} have shown that more than 80\% of the DRAM chips in the market are vulnerable to the Rowhammer attack including DDR4 chips having Target Row Refresh (TRR) mitigation. \cite{hassan2021uncovering} proposed a methodology that results in bit flips in 99.9\% of all DRAM rows on DDR4 chips with TRR protection. The Error Correcting Codes (ECC) mitigation has also been bypassed in \cite{cojocar2019ecc}. Rowhammer is a significant threat to shared cloud environments \cite{cojocar2020we, xiao2016one} as it can be launched across virtual machine (VM) boundaries and even remotely through JavaScript. Two research teams concurrently \cite{tatar2018throwhammer, lipp2020nethammer} showed even a remote machine can induce Rowhammer bit flips by sending network packets. More recently, \cite{half-double} have shown a combined effect of more than two aggressor rows to induce bit flips in recent generations of DRAM chips. All existing Rowhammer defenses including TRR, ECC, detection using Hardware Performance Counters, and changing the refresh rate can not fully prevent the Rowhammer attack \cite{gruss2018another, frigo2020trrespass}. The only requirement of the Rowhammer attack is that the attacker and the victim share the same DRAM chip, vulnerable to the Rowhammer attack. 

\newhl{Terminal Brain Damage~\cite{hong2019terminal} attack showed that DNN model weights are vulnerable to Rowhammer since bit-flip corruptions can alter the value of floating-point numbers significantly, causing accuracy degradation and even targeted misclassification. Deephammer~\cite{yao2020deephammer} showed that Rowhammer can deplete the accuracy of quantized DNN models as well. 
}

\subsection{Deep Neural Networks}
Deep Neural Networks (DNN) is a sub-field of Machine Learning, which are Artificial Neural Networks inspired by the biological neural cells of animal brains. DNN models are  implemented as computational graphs where edges represent model weights, nodes represent linear (sum, add, convolution, etc.), and non-linear operations (sigmoid, softmax, relu, etc.).  
DNN models are formed by multiple layers of weight parameters where each layer learns a different level of abstraction of the features hierarchically~\cite{zeiler2014visualizing}. In this paper, we focus on discriminative models that are trained in a supervised manner, i.e., with labeled data. Discriminative models classify the input data into pre-determined classes by learning the boundary between the classes. More formally, a DNN model $f$ is parameterized by $\theta$ maps the input samples $\{x_i\}$ into their corresponding classes $\{y_i\}$.
\paragraph{\textbf{Training}} 
The model parameters $\theta$ are optimized using the data pairs $\{x_i,y_i\}$ according to the following objective,

\begin{equation}
     \min_{\theta} F(\theta) \nonumber =  
    \sum_i \Big[\ell\Big(f(x_i, \theta), y_i\Big)\Big],
\end{equation}
where $F$ is the objective function, $\ell$ is a loss function, $\Delta\theta$ is the change in the model weights. The model is updated by backpropagating the errors through the layers~\cite{rumelhart1986learning}. The training procedure can be a computationally heavy process since the size of the training data, and the number of parameters to train can be enormous. Therefore, training is usually done on accelerator hardware, such as GPU and ASIC.
\paragraph{\textbf{Inference}}  
After the model weights reach an acceptable performance on the training data set, they can be deployed as a part of the service. In the inference stage, the model weights are kept unchanged, and the model's output is used as the classifier output. Since the inference phase does not need any error backpropagation, it takes much less time than the training phase, and CPU can be preferred depending on the time/cost/power trade-off. 

\subsection{Backdoor Attacks on DNN Models}\label{sec:backdoor} The terms \textit{Backdoor} and \textit{Trojan} are used interchangeably by different communities. Here we use Backdoor for consistency. In DNN models, we define a \textit{Backdoor} as a hidden feature that causes a change in the behavior triggered only by a particular type of input. 
In the literature, backdooring is applied with either benevolent intents, such as watermarking the DNN models~\cite{adi2018watermarking,Shafieinejad2021watermarking}, or with malicious purposes~\cite{gu2017badnets,bai2021targeted,chen2021proflip,liu2017trojaning, bagdasaryan2020blind,clements2018hardware}, as a \textit{Trojan} to attack the models.

In this work, we focus on \textit{Backdoor} as a type of \textit{Trojan} exploited by an attacker to cause targeted misclassification. 
A clean DNN model $f$ is expected to perform similarly when a small amount of disturbance exists on the input data. Therefore, $f(x_i+\Delta x, \theta)=y_i$ if and only if $f(x_i,\theta)=y_i$,  where $\Delta x$ is a small disturbance on the input $x$. We say a DNN model $f$ has a \textit{backdoor} if $f(x_i,\theta)=y_i)$ and $f(x_i+\Delta x, \theta) = \Tilde{y}$. 

Earlier works~\cite{liu2017trojaning, gu2017badnets,bagdasaryan2020blind,clements2018hardware} demonstrated that backdoor attacks pose a threat to the DNN model supply chain. Specifically, DNN models can be~\textit{backdoored} during the training phase if the model training is wholly or partially (transfer learning) outsourced~\cite{gu2017badnets}.
Moreover, compromised model-training code can be an attack vector for backdoor attacks since it can train a backdoored model even if the model is trained with the local resources and clean training data set~\cite{bagdasaryan2020blind}.


\section{Threat Model}
\label{sec:threat_model}

\begin{figure}
    \centering
    \includegraphics[width=0.8\columnwidth,trim={2mm 5mm 5mm 0},clip]{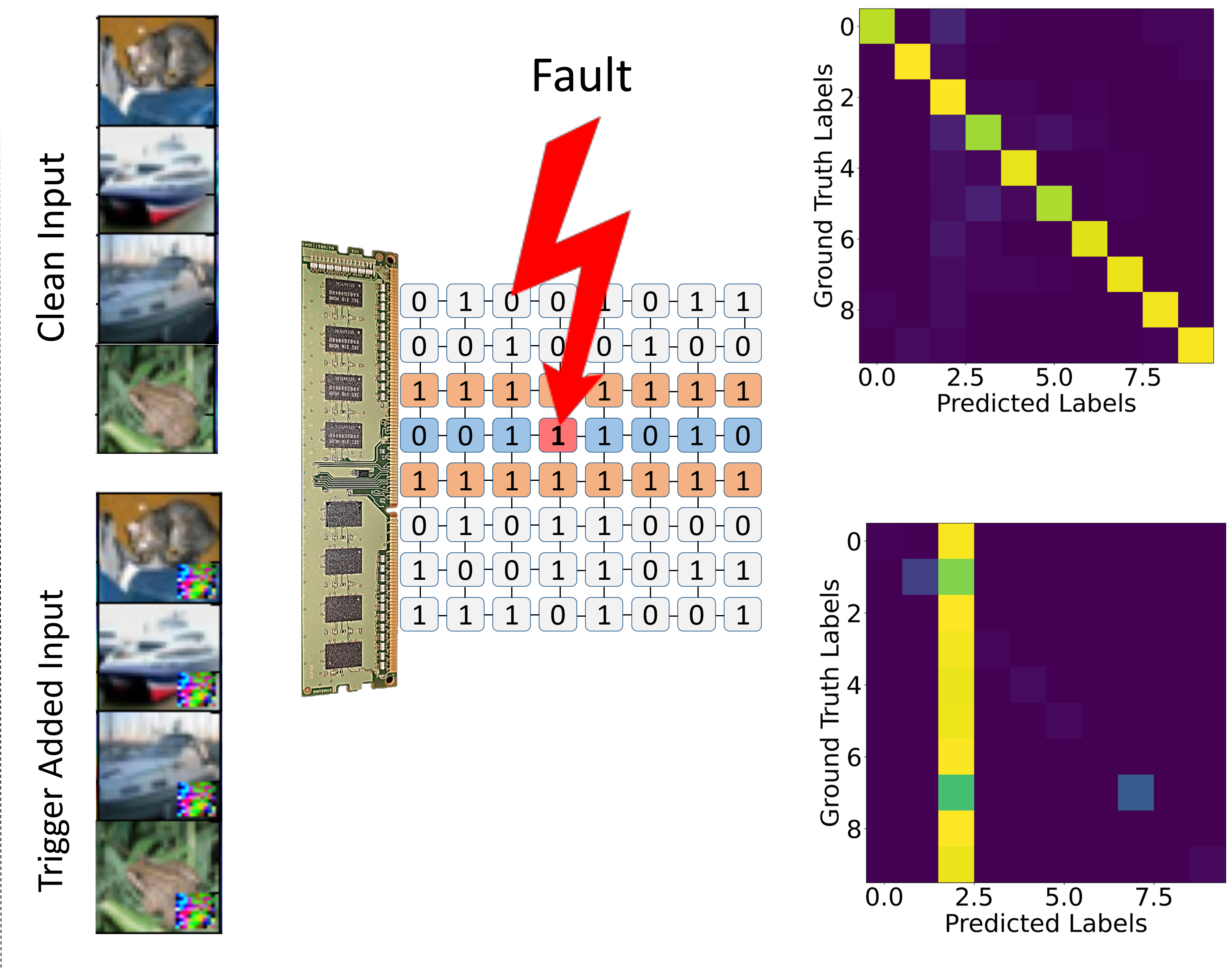}
    \caption{Backdoored Model behavior with clean inputs (top) and trigger added inputs (bottom). Fault injection to the model changes the behavior of the classifier, as shown by the confusion matrices.}
    \label{fig:attack_diagram}
\end{figure}

Same as in earlier works \cite{gu2017badnets,rakin2020tbt,liu2017trojaning,hong2019terminal,yao2020deephammer}, we assume that the attacker
\begin{itemize}
    \item knows the model architecture, parameters and the task of the target model;
    \item does not have access to the training hyperparameters or the training data set;
    \item has a small percentage of the unseen test data set;
    \item is involved only after the model deployment in a cloud server and therefore does not need to modify the software and hardware supply chain;
    \item resides in the same physical memory as the target model;
    \item has no more than regular user privileges (no root access).
\end{itemize}
Such threat models are well motivated in shared cloud instances targeting a co-located host running the model and in sandboxed browsers targeting a model residing in the memory of the host machine \cite{cojocar2020we, xiao2016one, de2021smash}. Moreover, the previous research on model stealing attacks~\cite{yu2020cloudleak,papernot2017practical,tramer2016stealing,correia2018copycat,juuti2019prada} validates our white-box attack assumption. The test data required by our attack does not belong to the victim and is not in the training data set. Hence, it can be easily collected and labeled by the attacker since the task of the target model is known.

To better understand our attack, we illustrate an example in Figure~\ref{fig:attack_diagram}. The attack works as follows:
\begin{enumerate}
    \item {\em Offline Phase - Profiling Target Model and Memory:} By studying the model parameters and the memory, the attacker generates a trigger pattern and determines the vulnerable bits in the target model. 
    \item {\em Online Phase - Rowhammer Attack:} After the target model is loaded into the memory, using Rowhammer, the attacker flips the target bits by only accessing its own data that resides in the neighboring rows of the weight matrices in the DRAM. 
    \item {\em Targeted Misclassification:} After the backdoor is inserted, the model will misclassify trigger-added input to the target class. The misclassification will persist until the backdoored model is unloaded from the memory. Since the model in persistent storage (or in the software distribution chain) is untouched, malicious modification to the model is harder to detect.
\end{enumerate}

\section{Backdoor Injection using Rowhammer}
\subsection{Offline Attack Phase}\label{subsec:offline}
In the offline phase of the attack, we optimize the trigger pattern and the bit-flip locations in the weight matrices. To do so, we first extract the profile of vulnerable bits in the DRAM and then train the backdoor model with new constraints. 

\subsubsection{\textbf{Memory Profiling For Adjacent Rows}}\label{sec:adjacentrows}
For the Rowhammer attack to work, we need to locate physical rows adjacent to victim rows that require finding physically contiguous memory addresses. We exploit SPOILER vulnerability~\cite{islam2019spoiler} in Intel processors to determine which virtual addresses within an array are contiguous physically.

After performing SPOILER and determining which addresses are contiguous physically, these addresses need to be filtered even further to addresses that are within the same bank. This is again performed using another timing side-channel attack known as row conflict \cite{pessl2016drama}, which measures the difference in read times between two addresses to determine if the row buffer for the bank was cleared, resulting in a longer read time and extrapolating bank continuity. 

\subsubsection{\textbf{Memory Profiling For Faults}}\label{sec:profiling}

Memory profiling is a process of finding vulnerable addresses in the DRAM. This process can be performed before the victim starts running. For DDR3 DRAMs, we implement a double-sided Rowhammer attack where we place a victim row between two attacker-owned rows. We set the victim rows to all zero and attacker rows to all one and repeatedly access the attacker rows. Then we check if there is any \textit{zero to one} flip in the victim row. We find the \textit{one to zero} flips similarly.   
For DDR4 systems, double-sided Rowhammer does not work due to the TRR mitigation implemented by the DRAM vendors. Therefore, we designate alternating rows to be attacker and victim.

Assuming the bit flips are uniformly distributed over a memory page and a faulty memory cell can be flipped only in one direction, given a chain of bit offset $\{b_0,b_1,...,b_{k+l-1}\}$ in a memory page, the conditional probability of finding a suitable target page $t$ in $N$ flippy pages can be calculated as 
\begin{multline}\label{eq:prob}
    p\big(t|\{b_{n_{0\rightarrow1}}\}\in\{0\rightarrow1\},\{b_{n_{1\rightarrow0}}\}\in\{1\rightarrow0\}\big) =\\
    1- \bigg(1- \prod_{i=0}^{k-1}{\dfrac{n_{0\rightarrow1}-i}{S-i}}\times\prod_{j=0}^{l-1}{\dfrac{n_{1\rightarrow0}-j}{S-k-j}}\bigg)^N,
\end{multline}
where "$n_{0\rightarrow1}$" and "$n_{1\rightarrow0}$" are the average numbers of faulty memory cells in a page, flippable in the direction from $0$ to $1$ and $1$ to $0$ respectively, which are device-dependent values, "$k$" and "$l$" are number of bit locations which need to be flipped in the direction from $0$ to $1$ and $1$ to $0$ respectively, and "$S$" is the total number of bits in a page. Previous research~\cite{mus2020quantumhammer} shows that $n_{0\rightarrow1}$" and "$n_{1\rightarrow0}$" are almost equal to each other. Therefore, Equation~\ref{eq:prob} can be reduced as,
\begin{multline}\label{eq:prob_reduced}
    p\big(t|\{b_{n_{0\rightarrow1}}\}\in\{0\rightarrow1\},\{b_{n_{1\rightarrow0}}\}\in\{1\rightarrow0\}\big) \approx\\
    1- \bigg(1- \prod_{i=0}^{k+l-1}{\dfrac{n_{0\rightarrow1}+n_{1\rightarrow0}-i}{S-i}}\bigg)^N.
\end{multline}

\begin{figure}
 \centering
      \begin{subfigure}[b]{0.40\textwidth}
         \centering
         \includegraphics[width=\textwidth]{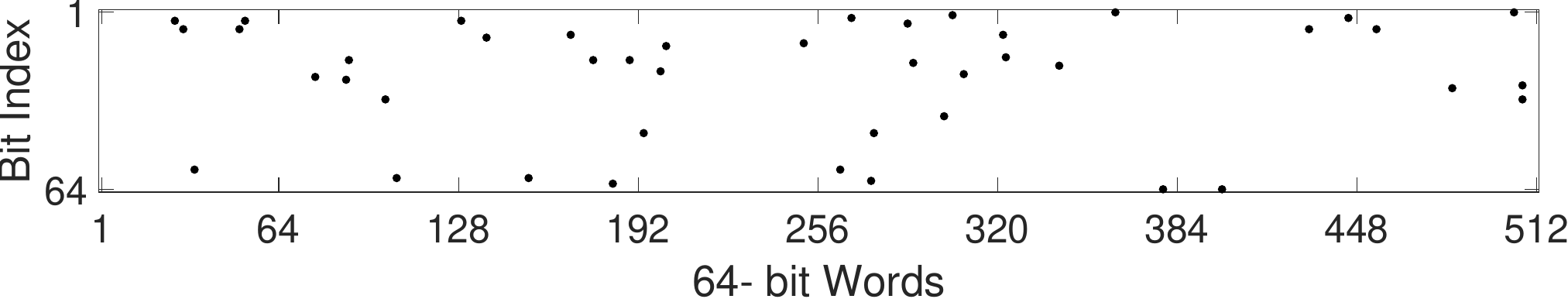}
         \caption{DDR3}
         \label{fig:DDR3}
     \end{subfigure}
     \hfill
     \begin{subfigure}[b]{0.40\textwidth}
         \centering
         \includegraphics[width=\textwidth]{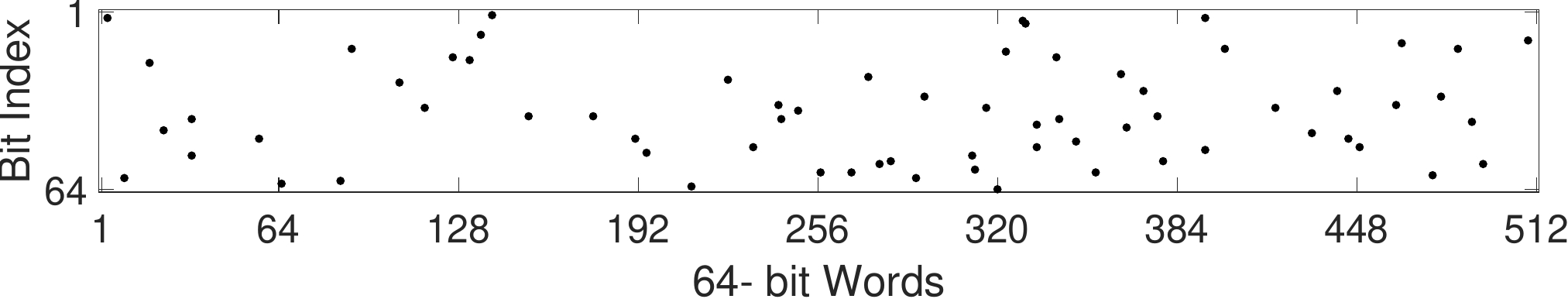}
         \caption{DDR4}
         \label{fig:DDR4}
     \end{subfigure}
          \caption{\label{fig:profile_page}The bit flip locations in the profiled 128MB memory buffer and one of the 4KB pages show the sparsity of the bit flips. Only about 0.036\% of the DRAM cells in the profiled memory are found to be vulnerable.}
\end{figure}

It takes 94 minutes to profile 128MB of memory, but this is done offline before the victim starts running. Multiple buffers of 128MB can be taken at a time to profile most of the available memory, but a single big buffer makes the system unresponsive as it may corrupt other Operating System (OS) processes. Figure \ref{fig:profile_page} shows the sparsity of the bit flips in the profiled 128MB buffer and one of the 4KB pages in DDR3 and DDR4 DRAM chips. 

Although we use state-of-the-art memory hammering techniques, we have found 34 bit flips in a 4KB page in DDR3. Overall, in the 128MB buffer, we have found 381,962 bit flips which are just \textbf{0.036\%} of the total cells in the buffer, as illustrated in Figure \ref{fig:profile_page}. 
For \hl{profiling} DDR4, we use a 15-sided Rowhammer attack. 
We tested 6 different DDR4 chips and averaged the number bit flips per page for each device. We also calculated the average number of bit flips per page for the memory profiles published by earlier work~\cite{tatar2018defeating} and summarized the results in Table~\ref{tab:flip_profiles}.

\begin{table}
\caption{Average number of bit flips per memory page for 14 DDR3 and 6 DDR4 chips. The tags in DRAM columns represent different brand/model information. The results for DDR3  results are calculated from double-sided Rowhammer profiles~\cite{tatar2018defeating}. DDR4 results are from the chips we profiled using n-sided Rowhammer. }
    \centering
    \begin{tabular}{c|c|c||c|c}
    \toprule
         & DRAM & \thead{Average \# of Flips\\ Per Page }& DRAM & \thead{Average \# of Flips\\ Per Page } \\
    \midrule
         \multirow{7}{*}{\rotatebox[origin=c]{0}{DDR3}} & A1 & 12.48 &  E1 & 12.46 \\
         & A2 & 1.92  &  E2 & 2.02 \\
         & A3 & 1.11  &  F1 & 28.77\\
         & A4 & 15.85 &  G1 & 1.62 \\
         & B1 & 1.05  &  H1 & 1.66 \\
         & C1 & 1.60  &  I1 & 8.28 \\
         & D1 & 1.08  &  J1 & 1.25 \\
    \bottomrule
        \multirow{2}{*}{\rotatebox[origin=c]{0}{DDR4}} 
         & K1 & 100.68 & L2 & 13.98\\
         & K2 & 109.48 & M1 & 2.04 \\
         & L1 & 3.12   & N1 & 2.72 \\
    \bottomrule
    \end{tabular}
        \label{tab:flip_profiles}
\end{table}

Specifically, we can estimate the probability of finding a suitable target page by fixing the DRAM-specific parameter $n_{0\rightarrow1}$ and $n_{1\rightarrow0}$  for a DRAM using Equation~\ref{eq:prob_reduced}. In line with the previous research~\cite{mus2020quantumhammer} we also observe that number of bit flips from $0$ to $1$ and $1$ to $0$ are almost equal. Therefore, using the results of our profiling experiments, we estimate that $n_{0\rightarrow1}+n_{1\rightarrow0}=34$. Total number of bits in a page is $S=32,768$, and the total number of pages is $N=32,768$ in a 128MB memory buffer where the page size is 4KB. Therefore, when $k=1$, i.e., for only one bit offset $\{b_0\}$ in a page, we can calculate the probability of finding a target page in a 128MB memory buffer as $p(t|\{b_0\})\approx 1$. Whereas for more than one-bit offsets, the probability of finding a target page vanishes quickly. Specifically, for  $\{b_0,b_1\}$, $p(t|\{b_0\}) = 0.03$ and for $p(t|\{b_0,b_1,b_2\}) = 0.00003$. Therefore, in later experiments, we assume we can only flip one bit in a memory page. 



\subsubsection{\textbf{Constrained Fine-Tuning with Bit Reduction (CFT+BR)}}\label{sec:cftbr}
We propose a novel {\em joint} learning framework based on constrained optimization to learn the bit flip pattern on the network weights as well as the data trigger pattern simultaneously. Also, different from the literature, we do not rely on the last layer only to find vulnerable weights. Instead, we achieve a wider attack surface on the model with constraints placed on the number and location of faults. 

To preserve the performance on clean data, given a collection of test samples $\{x_i\}$ and their corresponding class labels $\{y_i\}$, we propose optimizing the following objective:
\begin{multline}\label{eq:problem}
\min_{\Delta\theta\in\Delta\Theta}\max_{\|\Delta x\|_{\infty}\leq\epsilon} F(\Delta\theta, \Delta x)  = \\
    \sum_i \Big[(1-\alpha)\cdot\ell\Big(f(x_i, \theta+\Delta\theta), y_i\Big) \\  
      + \alpha\cdot\ell\Big(f(x_i+\Delta x, \theta+\Delta\theta), \Tilde{y} \Big)\Big],
\end{multline}
where $\Delta\theta$, $\Delta x$ denote the weight modification pattern and the data trigger pattern, $\Tilde{y}$ denotes the target label, $\ell$ denotes a loss function, $f$ denotes the network parameterized by $\theta$ originally, $\alpha\in[0,1]$ denotes a predefined trade-off parameter to balance the losses on clean data and triggered data. A large $\alpha$ value would cause the attack to give a more aggressive effort to increase the Attack Success Rate while sacrificing Test Accuracy, and a low $\alpha$ value would cause the attack to preserve the Test Accuracy while sacrificing the Attack Success Rate. Ideally, a moderate $\alpha$ value should be chosen to get a high Attack Success Rate while preserving the Test Accuracy as much as possible. Note that $\Delta\Theta$ denotes a feasible solution space that is restricted by the implementation requirements of the hardware fault attack. 

{\textbf{Rowhammer attack restriction in hardware:} allows realistically to flip only about one bit per memory page due to the physical constraints. Since the potentially vulnerable memory cells in the DRAM are sparse, the probability of finding a suitable target page to locate the victim is very low for more than one bit flip offsets (See Section~\ref{sec:profiling}). Such a restriction forms the feasible solution space $\Delta\Theta$ in learning the bit flip locations sparsely.}

To solve the constrained optimization problem defined in Equation~\ref{eq:problem}, we also propose a novel learning algorithm as listed in Algorithm~\ref{alg:offline} that consists of the following four steps:
\begin{algorithm}[t]
\caption{Learning realistic Rowhammer attack for hardware implementation}
\label{alg:offline}
\SetAlgoLined
 \KwIn{A DNN model with weights $\theta$, number of bits $N_{flip}$ that are allowed to be flipped in the memory, objective $F$, parameter $\epsilon$, learning rate $\eta$, and maximum number of iterations $T$}
 \KwOut{Backdoored model $\theta^*$ and trigger pattern $\Delta x^*$}
\BlankLine
$\Delta\theta^*\leftarrow\emptyset, \Delta x^*\leftarrow\emptyset$;

\For{$t\in[T]$}{
    \If{update the trigger == true}{$\Delta x^* \leftarrow \Delta x^* + \epsilon\cdot \sgn(\nabla_{\Delta x} F(\Delta\theta^*, \Delta x^*))$;}
    
    $\mathcal{M}\leftarrow Group\_Sort\_Select(|\nabla_{\Delta \theta} F(\Delta\theta^*, \Delta x^*)|,$ \\$
    N_{flip}, 'descending')$;
    
    $\Delta\theta^* \leftarrow \Delta\theta^* - \eta\cdot[\nabla_{\Delta\theta} F(\Delta\theta^*, \Delta x^*)]_{\mathcal{M}}$;
    
    \If{bit reduction == true}{$\theta^*\leftarrow\mathrm{Floor}((\theta+\Delta\theta^*)\oplus\theta) \oplus \theta$;}
}
\Return{$\theta^*, \Delta x^*$}
\end{algorithm}
\textbf{\textit{Step 1. Learning data trigger pattern $\Delta x$:}}
The goal of this step is to learn a trigger that can activate the neurons related to the target label $\Tilde{y}$ to fool the network. Trigger pattern generation starts with an initial trigger mask. Then, we use the Fast Gradient Sign Method (FGSM)~\cite{goodfellow2014explaining} 
to learn the trigger pattern. The update rule is defined as
\begin{equation}
    \Delta x = \Delta x^* + \epsilon\cdot \sgn(\nabla_{\Delta x} F(\Delta\theta^*, \Delta x^*)),
\label{eq:trigger}
\end{equation}
where $\Delta\theta^*, \Delta x^*$ denote the current solutions for the two variables, $\nabla$ denotes the gradient operator, and $\sgn$ denotes the signum function. $\epsilon\geq0$ denotes another predefined parameter to control the trigger pattern. Since it acts as a learning rate of the trigger, smaller values update the trigger slower but may be more effective in finding the optimal pattern.


\textit{\textbf{Step 2. Locating vulnerable weights:}} \label{sec:weight_search}
Now, given a number of bits that need to be flipped, $N_{flip}$, our algorithm learns which parameters are the most vulnerable. In this step, we apply two constraints to the optimization: 
\begin{itemize}
    \item C1. Locating one weight per bit flip towards minimizing our objective in Equation~\ref{eq:problem} significantly;
    \item C2. No co-occurrence in the same memory page among the flipped bits.
\end{itemize}


Recall that when a DNN model is fed into the memory, the network weights are loaded sequentially page-by-page, where each page is fixed-length and stored contiguously. We can view this procedure as loading a long vector by vectorizing the model. Therefore, to guarantee we choose at most one weight per memory page, we divide the network weight vector into $N_{flip}$ groups as equally as possible, as illustrated in Figure~\ref{fig:target_neurons}. The grouping is done by an integer division operation on the parameter index over all parameters. If the index of a parameter is $i_w$, the group ID of that parameter is determined as $i_w \mathbf{div}  (4096 * N_{group})$ where $N_{group}$ is the number of pages per bit flip, and $\mathbf{div}$ is integer division operation. $N_{group}$ depends on the chosen number of bit flips $N_{flip}$ and can be calculated as $N_{group} = N_w \mathbf{div} (4096*N_{flip})$ for a DNN model with number of parameters, $N_w$. After grouping the parameters, we rank the weights per group based on the absolute values in the gradient over $\Delta\theta$, \ie $|\nabla_{\Delta\theta}F|$ where $|\cdot|$ denotes the entry-wise absolute operator, in descending order. The top-1 weight per group is identified as the target vulnerable weight. 
Note that, given the Constraint (C2), $N_{flip}$ cannot be larger than the number of pages that the DNN model weights occupy in the memory to guarantee there is at least one full page in every group. The whole parameter selection process is represented with the following operation:
\begin{multline}
    \mathcal{M}\leftarrow Group\_Sort\_Select(|\nabla_{\Delta \theta} F(\Delta\theta^*, \Delta x^*)|, \\
    N_{flip}, 'descending'),
\end{multline}

\begin{figure}
    \centering
    \includegraphics[width=0.6\linewidth]{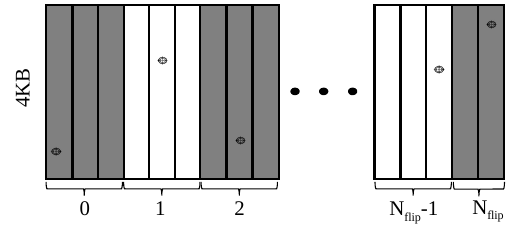}
    \caption{The illustration of targeted model weights across the DNN model weight pages in the memory. \bullsseye~denotes the targeted bit location in a page.
    }
    \label{fig:target_neurons}
\end{figure}



\textit{\textbf{Step 3. Adversarial fine-tuning}}
Now, given a collection of located vulnerable weights, denoted by $\mathcal{M}$, we only need to update these weights in backpropagation as follows:
\begin{equation}\label{eq:deltatheta}
    \Delta\theta = \Delta\theta^* - \eta\cdot[\nabla_{\Delta\theta} F(\Delta\theta^*, \Delta x^*)]_{\mathcal{M}},
\end{equation}
where $[\cdot]_{\mathcal{M}}$ denotes a masking function that returns the gradients for the weights in $\mathcal{M}$, otherwise 0's, and $\eta\geq0$ denotes a learning rate. 

\textit{\textbf{Step 4. Bit reduction }}
To meet the physical constraints of the Rowhammer, the final part of our attack procedure requires bit reduction. Rowhammer can only flip a very low number of bits in a 4KB memory page, and more than one faulty memory cell almost never coexists within a byte. 
Therefore, we define a bit reduction function as $\mathrm{Floor}(\theta \oplus \theta^*)$, where $\oplus$ denotes the bit-wise summation, and function $\mathrm{Floor}$ rounds down the number by keeping the most significant nonzero bit only. For instance, letting $\theta=1101_2$ and $\theta^*=1010_2$, then $\mathrm{Floor}(\theta \oplus \theta^*)=\mathrm{Floor}(0111_2) = 100_2$. In this way, we ensure that only one bit is modified in a selected weight while maintaining its change direction and amount as much as possible.

\subsection{Online Attack Phase: Flipping Bits in the Deployed Model}\label{subsec:online}
When we access a file from the secondary storage, it is first loaded into the DRAM and when we close the file, the OS does not delete the file from DRAM to make the subsequent access faster. 
If the file is modified, the OS sets the dirty bit of that modified page and writes back according to the configured policy. Otherwise, the file remains cached unless evicted by some other process or file. As Rowhammer is capable of flipping bits in DRAM, we can use it in the online attack phase to flip the weights of the DNN file as it is loaded in the page cache. The weight file is divided into pages and stored in the page cache. 
We can flip our target bits as identified by the backdoored parameters $\theta^{*}$, in Section \ref{subsec:offline}. The OS does not detect this change as it is directly made in hardware by a completely isolated process, and it keeps providing the page cached modified copy to the victim on subsequent accesses. Thus, the attack remains stealthy. In the online phase, we need to flip bits in the weight file in the required pages and page offsets. \hl{We achieve this in three main steps.}

\subsubsection{\textbf{\hl{Releasing the Flippy Rows}}} \hl{Flipping targeted bits in the model weights} requires manipulating the memory mapping of the weight file and placing the target pages to previously found flippy physical addresses. To control the memory mapping, we exploit the \textit{per-CPU page frame cache}. Page frame cache is an optimization implemented in the Linux kernel to utilize hardware caches better in the local CPU by reallocating the recently unmapped page frames in first-in-last-out order~\cite{bovet2005understanding}. As earlier works showed~\cite{chakraborty2020explframe, kwong2020rambleed,yao2020deephammer}, an attacker process can reliably map the victim page to the recently unmapped pages by exploiting the page frame cache. This unmapping-remapping process is shown by the pseudo-code in Listing \ref{lst:bait_pseudo}.
Although we do not need bit flips in all pages of the weight file, we need the target pages to be mapped to previously determined flippy page locations. 
\hl{We use a \texttt{buffer} with size $\texttt{baitPages} \times {PAGESIZE}$ to make sure the parameters we do not target in the weight file are not mapped into the flippy locations. The number of flippy pages and \texttt{baitPages} should sum up to the total number of memory pages consumed by the weight file.}

\begin{lstlisting}[ float=tp,frame=single,
                    label={lst:bait_pseudo},
                    caption= Pseudo-code showing how pages can be forced into a specific area in memory]
buffer = mmap(baitPages * PAGESIZE)
munmap(flippyPageAddr, PAGESIZE)
for(i = 0; i < bait_pages; i++)
    munmap(&buffer[i*PAGESIZE], PAGESIZE)

\end{lstlisting}

We match the target pages in the weight file to the flippy locations and the remaining pages to the non-flippy locations in our buffer. After obtaining a one-to-one mapping between the weight file and our buffer, we start unmapping in the reverse direction to fill the page frame cache. 

\subsubsection{\textbf{\hl{Mapping the Model Weights to Flippy Rows}}} 

\hl{After releasing the flippy pages and \texttt{buffer},} we immediately map the whole weight file from start to end \hl{using \texttt{mmap} function}.
The OS automatically maps the weight file to the unmapped locations in the buffer in the right order. An example case is shown in Figure~\ref{fig:memmassage} for a quantized ResNet20 model. Since all physical addresses match with the released pages of our buffer, there is a one-to-one mapping.

\begin{figure}[h]
    \centering
    \includegraphics[width=0.7\linewidth]{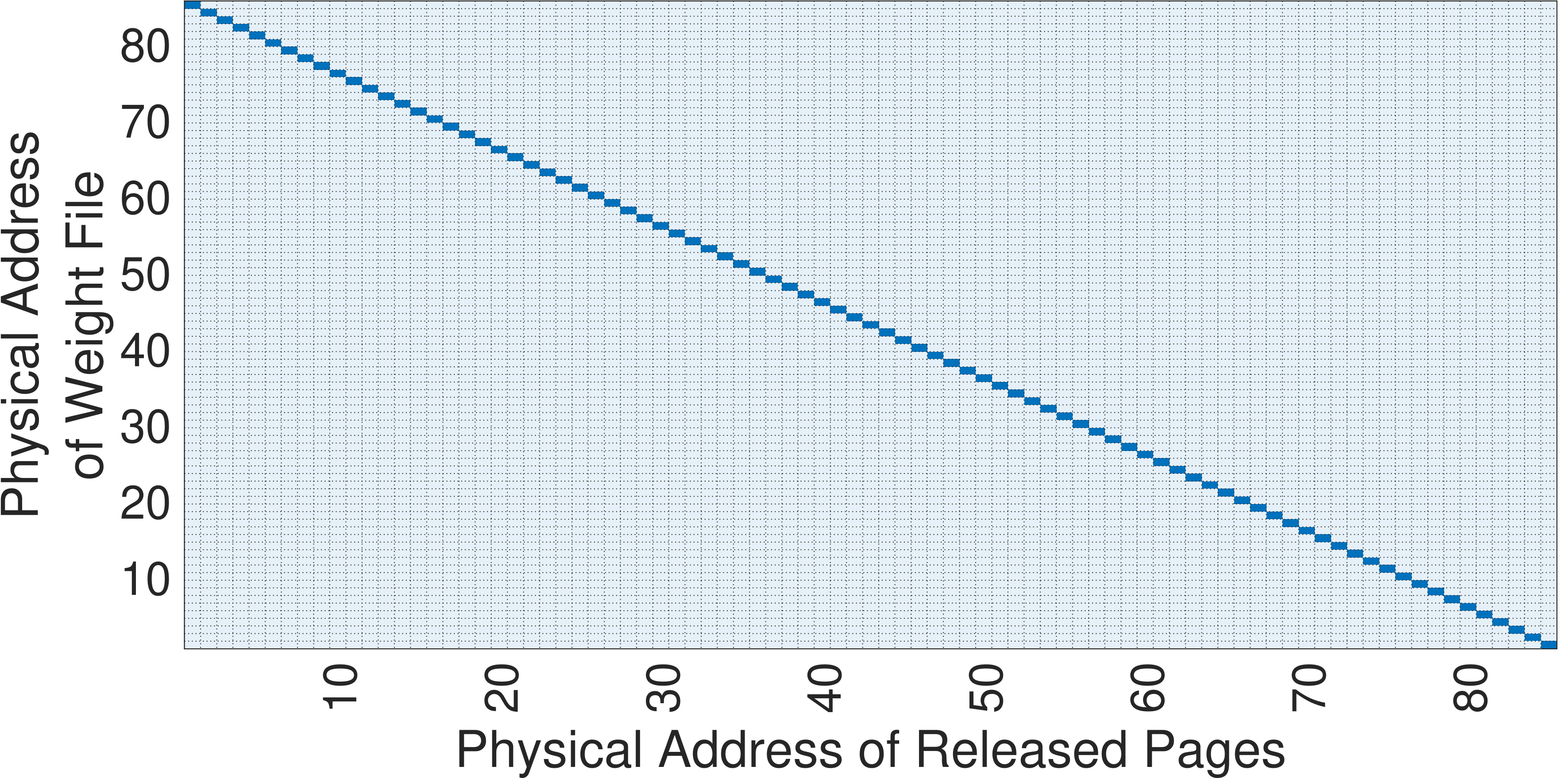}
    \caption{Physical Address of released pages vs ResNet20 weight file. First pages of the weight file are mapped to the last released pages of our buffer.}
    \label{fig:memmassage}
\end{figure}

\hl{Another way} to bring only the target pages of the weight file to the memory is by stating the file offset in the \texttt{mmap} function and using \texttt{fadvise} with \texttt{FADV\_RANDOM} flag to prevent the neighboring pages of the file prefetched by the OS, as proposed in~\cite{yao2020deephammer}. However, in our experiments, we observe that using \texttt{fadvise} does not reliably prevent prefetching. 

\subsubsection{\textbf{\hl{Flipping Bits in the Weight File}}}
Finally, the attacker rows are accessed repeatedly to flip bits at the same offsets as found in the offline phase but this time on the weight file. \hl{In our experiments, we use n-sided Rowhammer pattern~\cite{frigo2020trrespass} with 7 aggressor rows on DDR4 systems to bypass TRR protection and reproduce the bit flips found in the offline phase. Note that additional bit flips can occur if more than one bit flip is found within a single page. We evaluate the effect of these additional bit flips in Section~\ref{sec:evaluation}.}

\hl{After completing all the steps in Online Phase, the corrupted weights stay in the memory, and the attacker is able to add the pattern generated in Offline Phase to any image to trigger the backdoor and misclassify the input in a targeted way.}

\subsection{Weight Quantization}
The weights are stored as $N_q$-bit quantized values in the memory as implemented in NVIDIA TensorRT~\cite{migac2017tensorrt}, a high-performance DNN optimizer for deployment that utilizes quantized weights~\cite{nvidiatensorrt}. Essentially, a floating-point weight matrix $W_{fp}$ is re-encoded into $N_q$-bit signed integer matrix $W_q$ as $W_q = \mathrm{round}(W_{fp}/ \Delta w)$ where $\Delta w = \max(W_{fp})/(2^{N_q-1}-1)$. In our experiments, weights are 8-bit quantized and stored in two's complement forms.

\section{Evaluation}\label{sec:evaluation}

\subsection{Experimental Setup}\label{sec:expsetup}
To demonstrate the viability of our attack in the real world, we implemented it on an 8-bit quantized ResNet-18 model trained on CIFAR-10 using PyTorch v1.8.1 library. The clean model weights that are trained on CIFAR-10 are taken from~\cite{rakin2020tbt} for ResNet-18 and from~\cite{Idelbayev18a}(580 stars on GitHub) for other ResNet models. Moreover, we experimented on larger versions of ResNet models, such as ResNet50, trained on the ImageNet data set. For the models trained on ImageNet, we use pre-trained models of Torchvision library (9.1K stars on GitHub), which has been downloaded 28 million times until now~\cite{torchvision}.
We run the offline phase of our attack on NVIDIA GeForce GTX 1080Ti GPU and Intel Core i9-7900X CPU. Rowhammer experiments are implemented on DDR3 DRAM of size 2 GB (M378B5773DH0-CH9) and DDR4 DRAM of size 16 GB (CMU64GX4M4C3200C16).
\hl{The online phase experiments are conducted on a system running Ubuntu 20.04.01 LTS with a 5.15.0-58-generic Linux kernel installed, using a DDR4 DIMM with part number CMU64GX4M4C3200C16.} The inference is done on an Intel Core i9-9900K CPU with a Coffee Lake microarchitecture. DRAM row
refresh period is kept at 64ms which is the default value in
most systems. \hl{We use 7-sided Rowhammer to flip bits in the memory. We will provide an explanation for how we decide the number of aggressor rows in Section~\ref{sec:rowhammer_results_online}.}

We compare our approach with BadNet~\cite{gu2017badnets}, and TBT~\cite{rakin2020tbt} as well as fine-tuning (FT) the last layer. We also include the output of our Constrained Fine Tuning (CFT) without bit reduction in Table~\ref{tab:comparison} for comparison. We selected the baseline methods with the aim of creating a backdoor-injected model. We excluded the non-backdoor attacks, such as Deephammer~\cite{yao2020deephammer}, and Terminal Brain Damage~\cite{hong2019terminal}, in the performance comparison since they only aim to degrade the accuracy of the model. In contrast, we aim to keep the accuracy as high as possible while increasing the Attack Success Rate. For the offline phase results,
we keep all the bit flips in the weight parameters assuming they are all viable. In the online phase results, we keep the bits that are possible to be flipped by Rowhammer and exclude the others.
We use 128 images from the unseen test data set for all the experiments in CIFAR-10. TA and ASR metrics are calculated on an unseen test data set of 10K images. In all experiments, we used $\alpha=0.5$ for Algorithm~\ref{alg:offline}. The trigger masks are initialized as black square on the bottom right corner of the clean images with sizes 10x10 and 73x73 on CIFAR-10 and ImageNet, respectively. $\epsilon$ in Equation~\ref{eq:trigger} is chosen as 0.001.
For the ImageNet experiments, we use 1024 images from the unseen test data set to cover all 1000 classes. TA and ASR metrics are calculated on unseen test data set of 50K.

\subsection{Evaluation Metrics} \label{sec:metrics}

\textit{\textbf{Number of Bit Flips ($N_{flip}$):}}
As in \cite{yao2020deephammer,hong2019terminal,rakin2020tbt,bai2021targeted}, the first metric we use to evaluate our method is $N_{flip}$, which indicates how many bits are flipped in the new version of the model. The $N_{flip}$ has to be as low as possible because only a limited number of bit locations are vulnerable to the Rowhammer attack in DRAM. As the $N_{flip}$ increases, the probability of finding a right match of vulnerable bit offsets decreases. $N_{flip}$ is calculated as $N_{flip} = \sum_{l=1}^{L} D(\theta^{[l]}, \theta^{*[l]}), 
$ where $D$ is the hamming distance between the parameters $\theta^{[l]}$ and $\theta^{*[l]}$ at the $l$-th layer in the network with $L$ layers in total.


\textit{\textbf{DRAM Match Rate ($r_{match}$):}}
After a Rowhammer-specific bit-search method runs, the outputs are given as the locations of target bits in a DNN model. However, not all of the bit locations are flippable in the DRAM. Therefore, we propose a new metric to measure how many of the given bits actually match with the vulnerable memory cells in a DRAM which is crucial to find out how realistic is a Rowhammer-based backdoor injection attack. $r_{match}$ is calculated as,
    $r_{match} = \frac{n_{match}}{N_{flip}}\times (1-\frac{\delta}{S}) \times100$
where $n_{match}$ is the number of matching bit flips, $N_{flip}$ is the total number of bit flips, $S$ is the number of bits in a page, and $\delta$ is the number of accidental bit flips within a page. Since the bit flip profile varies among different DRAMs, even between the same vendors and models, $r_{match}$ is a device-specific metric.

\textit{\textbf{Test Accuracy (TA):}} In order to evaluate the effect of backdoor injection to the main task performance we use Test Accuracy as one of the metrics. Test Accuracy is defined as the ratio of correct classifications on the test data set with no backdoor trigger added. Ideally, we expect the backdoor injection methods to cause minimal to no degradation in the Test Accuracy in the target DNN models.

\textit{\textbf{Attack Success Rate (ASR):}}
We define the Attack Success Rate as the ratio of misclassifications on the test data set to the target class when the backdoor trigger is added to the samples. Attack Success Rate indicates how successful a backdoor attack is on an unseen data set.

\subsection{Rowhammer Attack on Deployed Model - Online}
\label{sec:rowhammer_results_online}

We experiment the online phase of the attack on DDR3 and DDR4 DRAM chips. We empirically observe that when there are multiple bits required to be flipped on the same 4KB page in a particular direction ($\{0\rightarrow1\}$ or $\{1\rightarrow0\}$), there is no matching target page in the 128 MB Rowhammer profile. This observation shows that multiple bit flips at desired page offsets and bit-flip direction is an unrealistic assumption.
On the other hand, we observe that there is always a matching page in the profiled memory buffer with a bit flip in the desired location and flip direction if there is at most one bit flip in the memory page. This observation is consistent with our probability analysis in Section~\ref{sec:profiling}. Apart from the targeted bit flips, we observed that some DDR4 DRAMs with large average bit flips in a page give accidental bit flips in addition to the target offsets  which reduces the $r_{match}$. 

\paragraph*{Effect of Number of Attacker Rows on Bit flips}

The idea of a multi-sided Rowhammer attack is that instead of a single row above and below the victim row being read, another victim is created above the attacker row, and another attacker above the new victim a variable number of times. Figure~\ref{fig:flips_per_sides} shows how the number of attacker rows changes the bit flip rate. 
\begin{figure}
    \centering
    \includegraphics[width=\columnwidth]{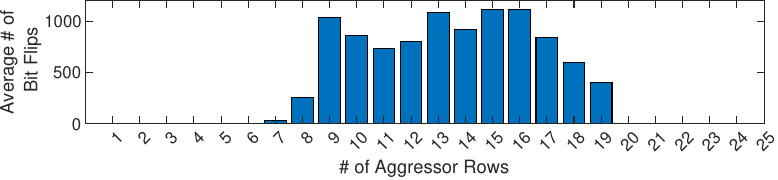}
    \caption{Average number of bit flips on an 8MB buffer vs the number of sides in an n-sided Rowhammer attack.}
    \label{fig:flips_per_sides}
\end{figure}

Figure~\ref{fig:turnoff_flips} shows that by reducing the number of aggressors in n-sided Rowhammer from 15 to 7, we can reduce the number of additional flips to ~4 bits per target page. \hl{Therefore, we use 7-sided Rowhammer in the later experiments.} Random bit flips outside the target location have a very limited effect on both TA and ASR since the target model weights are quantized~\cite{hong2019terminal}.

\begin{figure}[h]
    \centering
    \includegraphics[width=0.8\columnwidth]{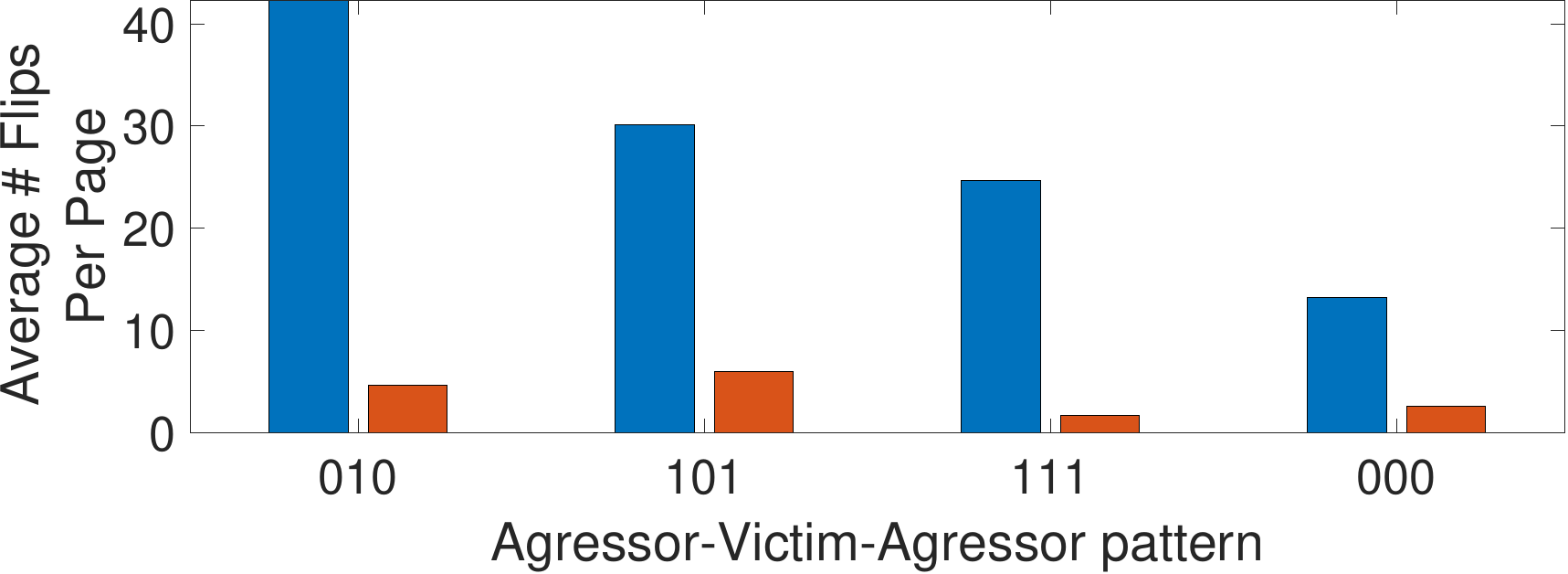}
    \caption{Average number of bit flips per page for 15-sided (blue) and  7-sided (red) Rowhammer attack patterns.}
    \label{fig:turnoff_flips}
\end{figure}

As shown in Table~\ref{tab:comparison}, we get $99.9\%$ $r_{match}$ for every DNN model we attack with CFT+BR since all of the required bit flips we need are in separate pages. Whereas BadNet, FT, TBT, and CFT, have very low numbers achieving as low as 1 bit flip since they require multiple-bit flips with specific locations and flip directions in the same memory page.

\begin{table*}[t]

    \caption{Comparison of our methods CFT, CFT+BR with the baseline methods BadNet, FT, and TBT on CIFAR10~\cite{krizhevsky2009learning} with ResNet-20/32/18, and ImageNet~\cite{deng2009imagenet} with ResNet-34/50. Our proposed CFT+BR results are written in bold. Note that the percentage of the backdoor parameter bits ($\Delta\theta$) that are actually flippable, $r_{match}$, must be near 100\% for a viable backdoor injection attack using Rowhammer.}
    \centering
    \begin{tabular}{c c c | c c c | c c c | c }
    \toprule
       &  &  &  \multicolumn{3}{c}{Offline Phase}  & \multicolumn{3}{c}{Online Phase} & \\
      Dataset   & \shortstack{Net} &  Method &	$N_{flip}$	  & TA(\%) & ASR(\%) & $N_{flip}$	  & TA(\%) & ASR(\%) & $r_{match}$(\%) \\
    \hline

                  \multirow{15}{*}{\rotatebox[origin=c]{0}{CIFAR10}}
                 & \multirow{5}{*}{\rotatebox[origin=c]{0}{\shortstack{ResNet20\\Acc: 91.78\%\\\#Bits: 2.2M\\\#Pages: 69}}}  &     BadNet       &    172,891      & 86.96   & 99.98  & 33 & 91.76   & 2.63  & 0.02 \\ 
                 & &       FT     & 2,238          & 84.36   & 97.10  & 1 & 91.72 & 2.90 & 0.04 \\ 
                 & &     TBT       & 44          & 86.61 & 95.43  & 1 & 91.72 & 4.71 & 2.27 \\ 
                  &  &     CFT      &   22   & 90.09 & 99.55 &  5 & 91.79 & 14.40  &  22.73 \\ 
                 &  &	CFT+BR      & \textbf{10}        & \textbf{91.24}      & \textbf{94.62} & \textbf{10}        & \textbf{89.04}      & \textbf{92.67}              &   \textbf{99.99} \\ 
        \addlinespace 
                  & \multirow{5}{*}{\rotatebox[origin=c]{0}{\shortstack{ResNet32\\Acc: 92.62\%\\\#Bits: 3.7M\\\#Pages: 116}}} &     BadNet       & 246,004          & 88.60   & 99.99 & 53 & 92.61  & 7.32  & 0.02 \\ 
                  & &      FT      & 2318          & 81.87   & 90.59  & 1 & 92.65 & 8.57 & 0.04 \\ 
                 & &     TBT       & 210          & 81.90  & 89.66  & 1& 92.66 & 8.42 & 0.48\\ 
                 &  &     CFT      &   39           &  90.25	 & 98.75 &  10	 &92.41 &  20.22	 & 25.64	  \\ 
                 &  &	CFT+BR      & \textbf{95}         & \textbf{91.77}       & \textbf{91.46} & \textbf{95}     & \textbf{89.56}       & \textbf{89.58} & \textbf{99.99}  \\ 
     \addlinespace 
                & \multirow{5}{*}{\rotatebox[origin=c]{0}{\shortstack{ResNet18\\Acc: 93.10\%\\\#Bits: 88M\\\#Pages: 2750}}} &     BadNet       & 1,493,301          & 87.61   & 99.88  & 416 & 93.06   & 12.45   &  0.03 \\ 
                 & &      FT      & 8,667          & 88.80   & 95.34  & 1 & 92.20   & 34.16   &  0.01\\ 
                 & &     TBT       & 95         & 82.87 & 88.82  &  1 & 92.60 & 48.12 & 1.05 \\ 
                 &  &     CFT      &   42          &  92.39& 99.90  & 11 & 91.52   & 0.36  & 26.19\\ 
                 &  &   CFT+BR        & \textbf{99}         & \textbf{92.95}   & \textbf{95.26} & \textbf{99}         & \textbf{90.71}   & \textbf{93.30}  &  \textbf{99.99}\\ 
     
      \midrule
      
   \multirow{10}{*}{\rotatebox[origin=c]{0}{ImageNet}}
                 & \multirow{5}{*}{\rotatebox[origin=c]{0}{\shortstack{ResNet34\\Acc: 73.31\%\\\#Bits: 172M\\\#Pages: 5375}}}  &      BadNet      & 441,047         & 70.81   & 99.73  & 100&  70.39 & 0.009 & 0.02\\ 
                 & &     FT       & 54,726         & 68.30   & 99.14  & 11 & 70.95  & 0.18  & 0.02\\ 
                 & &     TBT        & 553          & 72.69   & 99.86  & 1 & 70.97   & 0.05  & 0.18\\
                &  &     CFT       & 1509          & 70.25   & 99.76  & 388 & 69.93 & 0.10  & 25.71\\
                 &  &	CFT+BR        & \textbf{1463}   & \textbf{70.28}          & \textbf{72.92}   & \textbf{1463}   & \textbf{68.59}          & \textbf{71.42}  & \textbf{99.99}\\
        \addlinespace 
                  & \multirow{5}{*}{\rotatebox[origin=c]{0}{\shortstack{ResNet50\\Acc: 76.13\%\\\#Bits: 184M\\\#Pages: 5750}}} &     BadNet       & 359,516          & 73.98  & 99.11  & 129 & 66.43   & 0.05  & 0.04 \\ 
                  & &     FT       & 93,778         & 68.43  & 96.52  & 12 & 73.77   & 0.09 & 0.01 \\ 
                 & &     TBT       & 543          & 75.60   & 99.98  & 1 & 73.78   & 0.10 & 0.18 \\
                  &  &     CFT      & 1562          & 70.58   & 99.99  & 391  & 66.71   & 4.92 & 25.03\\
                 &  &	CFT+BR       & \textbf{1475}         & \textbf{70.64}   & \textbf{98.22}  & \textbf{1475}         & \textbf{68.94}   & \textbf{96.20}   & \textbf{99.99}\\
\bottomrule
    \end{tabular}
    \label{tab:comparison}
\end{table*}

\subsection{CIFAR-10 Experiments}
We experiment with our proposed method on ResNet18, ResNet20, and ResNet32 trained on CIFAR-10 along with the baseline methods, such as BadNet, FT, and TBT. We also compare our partial method, CFT, with our complete method (CFT+BR) which includes the \textit{Bit Reduction}. 
During the iterations of CFT+BR, we observed that the total loss spikes after each \textit{Bit Reduction} and quickly decreases again and eventually converges to a solution $\theta + \Delta\theta$ as described in Equation~\ref{eq:deltatheta}. Figure~\ref{fig:loss} shows the loss progress after each epoch with one batch of data while optimizing a constrained weight perturbation $\Delta\theta$ to a ResNet18 model on the CIFAR-10 data set. After every 100 iterations, we apply \textit{Bit Reduction}, which causes spikes in the loss curve.
\begin{figure}
    \centering
    \includegraphics[width=0.35\textwidth]{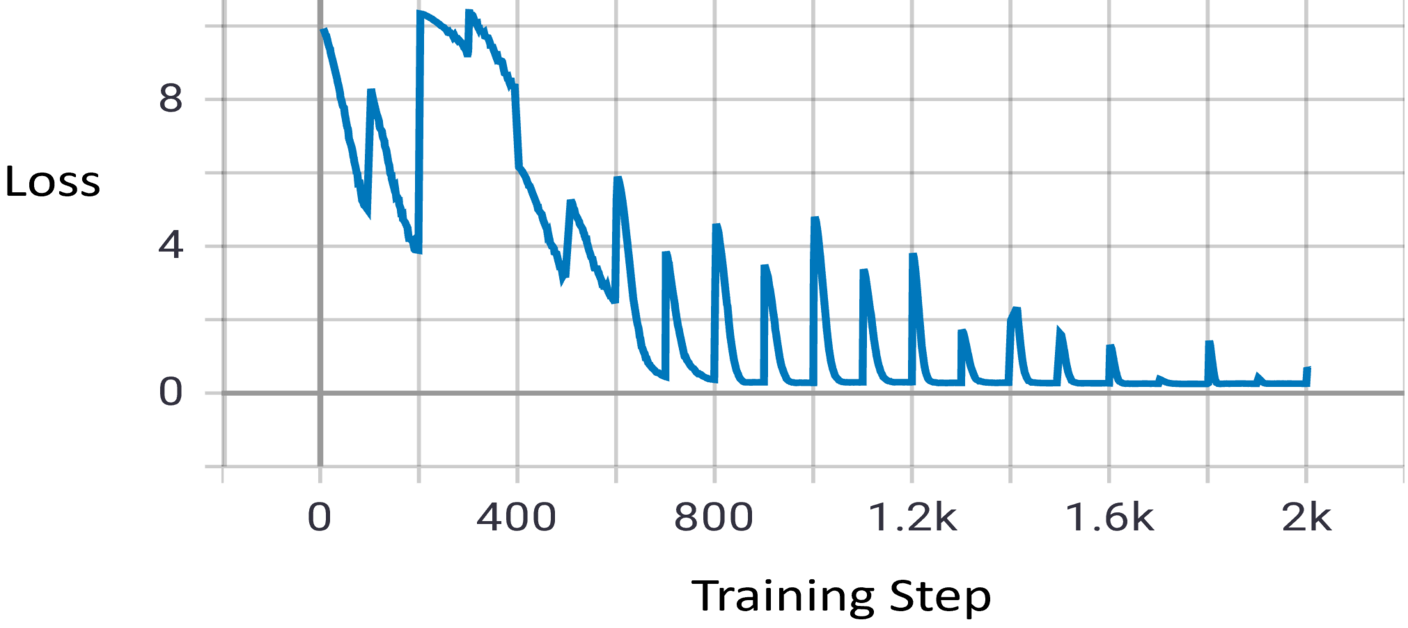}
    \caption{Total loss graph at every training iteration during the backdoor injection to the ResNet18}
    \label{fig:loss}
\end{figure}
We compare our method with baselines for both phases since our attack scenario includes offline and online phases. Recall that in the offline phase, the optimization takes place to find the vulnerable bit locations and generate a trigger pattern. First, we evaluate the modified models with the corresponding trigger patterns. Then, for each modified part of the weight parameters, we look for a matching target page location on the profiled memory, which constitutes the online phase. If multiple bits need to be flipped in the memory, we choose the one with the largest gradient value so that we get the maximum possible performance from the baselines. Finally, DRAM Match Rate $r_{match}$ is calculated as explained in Section~\ref{sec:metrics}. The experiment results are summarized in Table~\ref{tab:comparison}. 

BadNet and FT have no control over the $N_{flip}$ since they do not introduce any constraints during the optimization. Therefore, in the offline phase, BadNet requires up to one and a half million bit flips to inject a backdoor successfully. Although FT modifies only the last layer while keeping the other layers constant, meaning fewer bit flips than BadNet, we observe that up to 8,667 bits have to be flipped.
TBT has control on the number of modified parameters which enables partial control on the $N_{flip}$ since the number of modified parameters limits the maximum value $N_{flip}$ can get. Therefore, we select the results that reproduce their claimed performance in the original work~\cite{rakin2020tbt} without modifying too many weight parameters and increasing the $N_{flip}$ too much, and thus, decreasing $r_{match}$ further. In the offline phase, TBT finds a much smaller number of bits compared to BadNet and FT due to the limit on the modified parameters.
Our experiments show that the CFT+BR method successfully injects a backdoor into ResNet20 model with \textbf{91.24\%} TA and \textbf{94.62\%} ASR by flipping only \textbf{10 bits} out of 2.2 million bits in the DRAM.
In ResNet32 and ResNet18, CFT+BR achieves 91.46\% and 95.26\% ASR, respectively, with a maximum of 1.66\% degradation in the TA. 
We observe that $N_{flip}$ values in BadNet and FT depend heavily on mode size. As the total number of bits increases, they require more bit-flips to achieve similar performance. On the other hand, we do not observe a significant dependence on the model size in TBT, CFT, and CFT+BR methods in terms of $N_{flip}$, TA, and ASR.
In BadNet, FT, and TBT, the bit flips are concentrated within the same pages. Especially FT and TBT targets on the last layer of the DNN models. Since the last layer of the Resnet20, ResNet32, and ResNet18 models occupy only one memory page in DRAM, the bit-flip locations found in the offline phase of FT and TBT reside within a single page. For instance, 210 bit-flips found by TBT on ResNet32 are all on the same page. However, as we mention in Section~\ref{sec:profiling}, only the pages with one targeted bit location can be found in DRAM in practice. 
Therefore, we choose the bit flip with the largest gradient in a memory page and keep it modified and return the other parameters to their original values. Finally, we evaluate their performance on the test data set. In the ResNet20 and ResNet32 models, we observe that the ASR of BadNet, FT, and TBT drops down below 10\% while the Test Accuracy values increase back to their original values. We claim that the significant decrease in ASR values can be explained by the diffusion effect of optimizing the parameters in an unconstrained way. 
When the attack is implemented on DRAM using Rowhammer, $r_{match}$ values of BadNet, FT, and TBT are lower than 3\% for every DNN model. In CFT, $r_{match}$ is relatively higher than the other baseline methods since it modifies only one parameter in a page. However, it does not put a constraint on the number of bit flips within a byte during the optimization. Therefore, the attack performance degrades drastically in practice. In all experiments, CFT+BR has 99.9\% $r_{match}$ since it already considers the bit locations that can be flipped during the attack. Since the bit flips are sparse across different memory pages in CFT+BR, \textbf{100\%} of the bit flips can actually be flipped. A small number of bits may be flipped in random locations, but it does not affect the performance of the attack significantly. We show that lower $r_{match}$ values lead to low ASR in backdoor injection attacks using Rowhammer.

\subsection{ImageNet Experiments}

We also compare our method with the baseline methods on models trained on the ImageNet ILSVRC2012 Development Kit~\cite{deng2009imagenet} data set, which consists of 1000 classes of visual objects. We used pre-trained ResNet34 and ResNet50 from the model zoo~\cite{torchvision} as the target models. ResNet34 and ResNet50 include 172 million and 184 million bits, respectively. Note that both the model and data set sizes are significantly larger compared to our CIFAR-10 experiments. As the TA and  ASR, we use top-1 accuracy results. The results are summarized in Table~\ref{tab:comparison}. The same comparison methods we apply in CIFAR-10 are valid in ImageNet experiments as well.

In the offline phase of the attacks, we observe that each method shows a different response to the increase in the model and data set sizes. For instance, BadNet and FT require more than 350K and 50K, respectively. Compared to CIFAR-10 models, BadNet is not affected significantly. However, $N_{flip}$ for FT becomes 17 times larger on average on the ImageNet models.
TBT locates around 550 $N_{flip}$ on the ResNet34 and ResNet50 models in the offline phase, which is 5 times larger on average than the CIFAR-10 experiments.
CFT and CFT+BR locate around 1500 $N_{flip}$ on the ResNet34 and ResNet50 models in the offline phase, meaning 45 times and 22 times larger for CFT and CFT+BR, respectively.

In the online phase, we observe that none of the baseline methods has a significant attack performance. For instance, in the BadNet method, although the model sizes increase ~5.5 times, the number of modified pages increases only ~1.5 times on average. Similarly, TBT modifies only one page in the last layer of the ResNet34 and ResNet50 models, even though the last layers of the models have more than 10 pages. This clearly shows that as the model size increases, the density of bit flips required by the baseline models increases, meaning the attack tends to focus on certain regions instead of uniformly distributing the bit flips. The high density of the bit flips leads to $r_{match}$ rates as low as 0.02\%. Although FT modifies most of the pages in the last layer, the fact that the bit locations are not optimized at the beginning causes vanishing ASR. 
Overall, we observe that the claimed ASRs can be achieved only when $r_{match}$ is large enough. Although CFT achieves much larger $r_{match}$ values than the other baseline methods, lacking \textit{Bit Reduction} makes the attack focus on multiple bit flips within 8-bit parameters, which, in return, causes lower than 5\% ASR on the models trained with ImageNet data set.
In contrast, CFT+BR can inject the backdoor to ResNet models with up to 96.2\% ASR and a maximum of 7.2\% degradation in the TA, which makes it the best-performing backdoor injection attack compared to the baseline methods. These results show that our approach generalizes well to larger data sets and models. Note that although $N_{flip}$ increases as the model gets larger in CFT+BR, it is still possible to flip these bits with 99.99\% $r_{match}$ due to the sparse distribution. 

\subsection{Generalization to Other DNN Architectures}\label{sec:othermodels}
We experiment on other DNN architectures, such as VGG11, 
 VGG16,
  to show that our attack generalizes. We show that CFT+BR can successfully locate vulnerable bits and achieves over 90\% Attack Success Rate in VGG architectures. The results are summarized in Table~\ref{tab:othermodels}.

\begin{table}[h]
\caption{CFT+BR experiment results on VGG architectures}
    \centering
    \begin{tabular}{c|c|c|c|c}
    
    \toprule
        Model & Base Acc & TA [\%] & ASR [\%] & $N_{flip}$ \\
    \midrule
        VGG11 & 92.35 & 92.70 & 100 & 30\\
        VGG16 & 92.68 & 92.57 & 90.85 & 100\\   
    \bottomrule
    \end{tabular}
 
    \label{tab:othermodels}
\end{table}

\section{Potential Countermeasures}
We analyze some of the prominent countermeasures proposed for mitigating bit-flip attacks against DNN models.

\subsection{Prevention-Based Countermeasures}

\begin{figure*}[t]
  \centering
      \centering
      \includegraphics[width=0.75\textwidth]{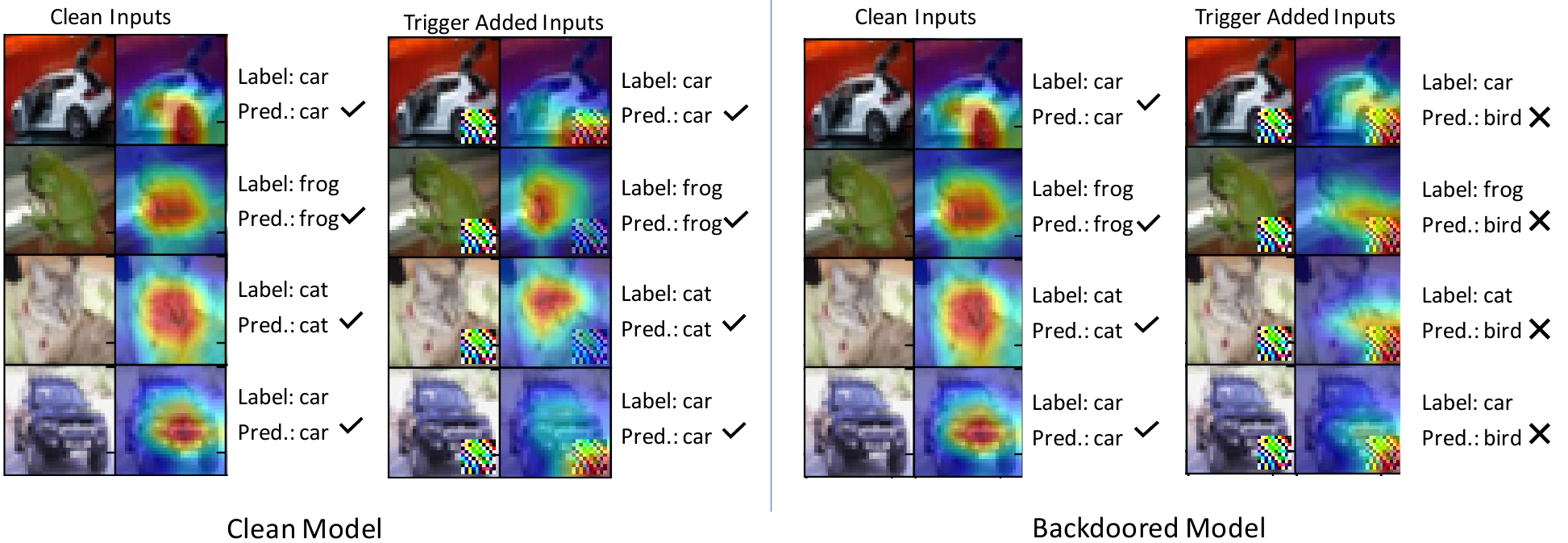}
\caption{The change in GradCAM~\cite{selvaraju2017grad} heatmaps that belong to ResNet18 before the attack (left) and after the attack (right). The focus of the model shifts through the trigger pattern if it is backdoored.}
      \label{fig:grad}
\end{figure*}

\textit{\textbf{Binarization-Aware Training~\cite{he2020defending}\footnote{Binarization-Aware Training and Piecewise Weight Clustering implementations are taken from \url{https://github.com/elliothe/BFA}.}}} is a method that uses \textit{Binarized Neural Networks (BNNs)}~\cite{hubara2016binarized, rastegari2016xnor} to increase the resistance of DNNs against the bit flip attacks. 
This method significantly reduces the network size.  For instance, a binarized ResNet-32 model occupies only 65 pages in the memory. Although 65 bit flips are not enough to inject a backdoor using Rowhammer, $N_{flip}$ cannot be larger than the number of pages occupied by the model. Therefore, our experiments show that using BNNs is an effective defense against our attack since it aggressively decreases the size of the network and, consequently, the maximum $N_{flip}$. However, reducing the model size causes accuracy degradation as a performance overhead. Note that BNNs may still be vulnerable to other fault attacks which do not require the same physical constraints, such as sparse faulty bit locations. 

\textit{\textbf{Piecewise Weight Clustering (PWC)~\cite{he2020defending}}} is a relaxation of BNNs. 
With PWC, an additional penalty term is introduced to the inference loss function, which forces model weight distribution to form two clusters.  We experiment with our attack against a ResNet32 model trained with PWC penalty term in the loss function. We observe a strengthened trade-off between the TA and ASR during the optimization.
For instance, the ASR drops down to 43.42\% when TA is 89.66\% with 112 $N_{flips}$. 
On the other hand, our attack achieves 98.49\% ASR while degrading the TA down to 9.9\% with the same $N_{flips}$. The results show that training the model with PWC does not protect against accuracy degradation and even targeted misclassification attacks. However, it makes it harder to inject stealthy backdoors. 

\subsection{Detection-Based Countermeasures}  Possible defense techniques focusing on detecting the attacks on the model weights~\cite{li2020deepdyve,liu2020concurrent,li2021radar,chou2020sentinet} come with an overhead because they need to be deployed together with the model into the machine learning product.

\textit{\textbf{DeepDyve~\cite{li2020deepdyve}}} is a dynamic verification method that uses a checker model along with the original model for mitigating the transient faults in the inference.
It assumes both models predict the same results for the same inputs most of the time. When the results of the two models are the same, the result is accepted immediately. However, if the results are different, the inference is repeated, and the second result from the original model is accepted. 
DeepDyve assumes the fault in the model is transient and does not appear in consecutive queries. However, the bit flips introduced by Rowhammer stay in the memory until being reloaded from the disk. Since the transient assumption does not hold, even if a checker model raises an alarm and repeats the inference, the new inference is made by the backdoor-injected model and will not be detected.

\textit{\textbf{Weight Encoding\cite{liu2020concurrent}}} proposes additional matrix multiplication and weight extraction. Thus, this method can detect only the topmost sensitive layers in the network to keep the overhead low. However, our attack can target all layers to inject a backdoor. Therefore, the spatial locality assumption does not hold with our attack. 
 Using the overhead numbers in~\cite{liu2020concurrent} for ResNet-34, we estimate the time and storage overhead against our attack. Since the time complexity of weight encoding $d_j = r(y_j), y_j=$ $\phi (\sum_{i=0}^{N-1} B_i \cdot K_{ij})$ is $O(N^2)$, where $B$ is $Z^N$, and $K$ is $R^{NxM}$, the estimated execution time overhead of the method is  834.27 seconds.  Since the storage complexity of the Weight Encoding is linear, the storage cost for ResNet34  is estimated as $(0.141 / 8192) \times 21779648 = 374.86 MB$, which is 446\% storage overhead, showing that the proposed method is not scalable.

\textit{\textbf{RADAR}~\cite{li2021radar}}  is a checksum-based detection method during inference. It divides the weights into groups and gets the checksum of the most significant bits of parameters in each group. The original checksum values of the parameters are stored along with the model and are validated with the original signatures at every inference time. 
The optimization constraints can be further increased to avoid flipping the MSB of the weight parameters in our attack, which can bypass the detection.
Assuming linear time complexity, time overhead goes up to 40.11\% for full-size bit protection in ResNet20.

\textit{\textbf{SentiNet~\cite{chou2020sentinet}}} filters the adversarial inputs using GradCAM heatmaps~\cite{selvaraju2017grad}. 
We use the GradCAM implementation from \cite{jacobgilpytorchcam} to analyze the output of four sample images that are labeled as \textit{car, frog, cat} and \textit{car} respectively (See Figure~\ref{fig:grad}.). Before the attack, the model correctly classifies all images with or without the trigger pattern. If the trigger pattern does not overlap with the major features in the image, e.g. \textit{frog} and \textit{cat}, the main focus of the model stays on the object. However, if the trigger pattern overlaps with the main features, e.g. the wheel of the \textit{car}, the focus is shifted towards the trigger pattern. 
After the attack, regardless of the trigger and object overlap, the focus of the model shifts towards the trigger pattern, and the model misclassifies all images to the target class, \textit{bird}.
Therefore, although a GradCAM-based approach can possibly filter the adversarial inputs, it will also produce false positives even if the model is clean and works correctly.

\subsection{Recovery-based Countermeasures} 
\textit{\textbf{Weight Reconstruction:}} Li et al.~\cite{li2020defending} propose \textit{Weight Reconstruction}\footnote{Weight Reconstruction implementation is taken from \url{https://github.com/zlijingtao/DAC20_reconstruction}.} to recover the clean network after a bit flip attack occurs. \textit{Weight Reconstruction} aims to recover from an accuracy degradation caused by the attack.
After a bit flip occurs in a weight parameter, the effect of the change is distributed onto other parameters to reduce the overall effect on the model performance. We experiment with our CFT+BR attack against a ResNet32 defended by Weight Reconstruction to evaluate the effectiveness of the proposed defense method. We applied our attack in two different scenarios. 
In the first scenario, the attacker is not aware that the model is defended by Weight Reconstruction and applies the offline phase of the attack 
as described in Section~\ref{sec:cftbr}. As a baseline, our attack achieves 91.46\% ASR and 97.77\% TA by flipping 95 bits in the memory. After applying Weight Reconstruction, we observed that ASR and TA become 32.89\% and 91.02, respectively. 
In the second scenario, the attacker is aware that the model is defended by Weight Reconstruction and applies the offline phase of the attack against a model with Weight Reconstruction. However, if the attacker is aware of the defense and applies CFT+BR on a defended model, our attack successfully bypasses Weight Reconstruction by achieving 94.04\% ASR and 89.51\% TA. Therefore, the Weight Reconstruction approach does not protect the models when the attacker knows the applied defense.


 \section{Related Works}\label{sec:relatedworks}
 \paragraph{Rowhammer Attacks on DNNs} \newhl{We compare our work with Terminal Brain Damage~\cite{hong2019terminal} and Deephammer~\cite{yao2020deephammer} in terms of the following factors:}
%

\newhl{\textit{Attacker's Objectives:} 
The main difference between our work and previous works is the goal of the attack. In both \cite{hong2019terminal} and \cite{yao2020deephammer}, the attacker's objective is to degrade the inference accuracy of the model on legitimate inputs and cause a denial of service. In contrast, our attack objective in this work is to keep the inference accuracy for legitimate inputs the same and misclassify all trigger-added inputs to a target class in stealth by using a unified objective function given in Equation~\ref{eq:problem}.}

\newhl{\textit{Assumptions:}
All \cite{hong2019terminal}, \cite{yao2020deephammer}, and our work assume the attack takes place in a cloud environment where the model is loaded into system's shared memory and stays unchanged. Unlike \cite{yao2020deephammer}, we do not assume the availability of huge page configuration to bypass virtual to physical translation.} 

\newhl{\textit{Attacker Capabilities:} 
Same as our attack, \cite{yao2020deephammer} and \cite{hong2019terminal} assume the attacker knows the model architecture and parameters. \cite{hong2019terminal} also considers black-box setting with random bit flips. Since our attack objective is more sophisticated, our attack is not applicable in a black-box setting.
}

\newhl{\textit{Attack Time:}
 \cite{yao2020deephammer} configures the hammering time for each row as 190ms. Since \cite{hong2019terminal} only simulates the attack, they assume it is 200ms in the calculations. In our setup, it takes 800ms to hammer one row using a 15-sided pattern during the profiling phase and 400ms using a 7-sided pattern during the online phase. Note that previous works consider only double-sided Rowhammer, which takes less time but is not effective on DDR4 chips with TRR mitigation. Total online attack time varies between different models and can be estimated by multiplying the hammering time by $N_{flip}$. 
}

\newhl{\textit{Stealthiness/Detectability:} 
Due to the difference in the attack objectives, the stealth of the attacks is also different.}
For instance, Test Accuracy after the attack on VGG16 is given as around 10\% in both \cite{hong2019terminal} and DeepHammer. However, we can preserve the Test Accuracy at over 92\% after our attack while being able to misclassify over 90\% of all instances with an attacker-generated trigger pattern. Since we can preserve the Test Accuracy close to the base accuracy of the models, our attack is stealthy.

\newhl{\textit{Comparison of Accuracy Degradation:}
Although the goal of backdoor injection is not accuracy degradation, the resulting degradation on trigger-added inputs is comparable to \cite{hong2019terminal} and \cite{yao2020deephammer}. In  VGG16 trained on CIFAR10, when we add trigger pattern to all images, we see the accuracy of the model to be 18\% (an 80\% relative accuracy degradation from baseline). Alternatively, \cite{yao2020deephammer} and \cite{hong2019terminal} claim relative accuracy degradations of VGG16 to be 88\% and 90\%, respectively (after the attack the models only produce a correct output 10\% of the time). 
}
%
\paragraph{Accuracy Degradation Attacks}
 Bit-Flip Attack~\cite{rakin2019bfa} degrades the accuracy of DNN models to random guess using a chain of bit flips. 
 Targeted Bit-Flip Attack~\cite{rakin2020tbfa} is shown to be capable of misclassifying the samples from single or multiple classes to a target class on quantized DNN models. Although these works show that DNN model performance can be damaged permanently by flipping a limited number of bits in the weight parameters, these attacks do not make use of an attacker-controlled backdoor trigger. Therefore, they have very limited control over stealthiness. 
A binary integer programming-based approach was proposed by Bai et al.~\cite{bai2021targeted} to find the minimum number of bit flips required to make the model misclassify a single image sample into a targeted class.
\paragraph{ML Backdoor Attacks}
Garg et al.~\cite{garg2020perturbation} observed that adversarial perturbations on the weight space of the trained models could potentially inject Backdoor, but it requires either social engineering or full privileged access to replace the target model with the backdoored model.
Recently, \cite{rakin2020tbt} and~\cite{chen2021proflip} showed that backdoor attacks could be implemented by changing only a small number of weight parameters.
However, both of the works assume any bit location in the memory can be flipped, which is not practical. Therefore, the practicality of software-based backdoor injection attacks during the inference phase is still an open question due to the practical constraints that have been overlooked in previous works.

\section{Discussion}\label{sec:discussion}
\textit{\textbf{Effect of Huge Pages:}} We assume huge pages are not available since they give an advantage for finding contiguous memory in physical address space. Even though the target model uses huge pages, the memory controller would still fragment the huge page into 8 KB rows in DRAM due to the fixed row size. Also, each chunk is mapped into different banks in order to increase parallel access. 
For example, if there are 64 banks in the system, a 2 MB huge page would be fragmented into 64 chunks and 4 neighbor rows in the DRAM. Although this may hurt the n-sided Rowhammer pattern, it would still be possible to sandwich each chunk and do Rowhammer. Note that, in memory systems with multiple DIMMs, and ranks, the number of banks also increases, which would decrease the size of the chunks down to a single row. In that case, a regular double-sided or n-sided Rowhammer attack would still work.
Since an attacker can choose to profile 4 KB pages in DRAM, finding 512-bit flips in 2 MB would still be practical.

\textit{\textbf{Application on Other Security Critical Tasks}}
The proposed attack method is a generic approach and agnostic
to the downstream tasks. Therefore, it would work on models used in other safety-critical tasks, such as voice recognition applications.

\section{Conclusion}
We analyzed the viability of a real-world DNN backdoor injection attack. Our backdoor attack scenario applies to deployed models by flipping a few bits in memory assisted by the Rowhammer. Our initial analysis performed on hardware showed that earlier proposals fall short in assuming a realistic fault injection model. We devised a new backdoor injection attack method that adopts a combination of trigger pattern generation and sparse and uniform weight optimization. In contrast to earlier proposals, our technique uses all layers and combines trigger pattern generation, target neuron selection, and fine-tuning model parameter weights in the same training loop. Since our approach targets the weight parameters uniformly, it is guaranteed that no more than one bit in a memory page is flipped. Further, we introduced new metrics to capture a realistic fault injection model. This new approach achieves a viable solution to target real-life deployments: on CIFAR10 (ResNet 18, 20, 32 models) and ImageNet (Resnet34 and 50 models) on real hardware by running the Rowhammer attack achieving Test Accuracy and Attack Success Rates as high as 92.95\% and 95.26\%, respectively. We also showed that our attack works on other architectures, such as VGG11 and VGG16.
Finally, we evaluated the prominent defense techniques against our backdoor injection attack. We concluded that the proposed countermeasures are either not effective or introduce significant overhead in terms of time and storage.

\section*{Acknowledgements}
We thank our anonymous reviewers for their insightful feedback. This work is supported by the National Science Foundation, under grants CNS-1814406, CNS-2026913, and CCF-2006738, and by the U.S. Department of State, Bureau of Educational and Cultural Affair’s Fulbright Program.

\bibliographystyle{IEEEtran}
\bibliography{references}

\appendix

\subsection{Probability Analysis}
\label{apx:prob}
We further analyze Equation~\ref{eq:prob_reduced} with the numbers calculated in the Table~\ref{tab:flip_profiles}. 

First, we calculate the probability of finding a target page t for each $N$ values and three different $k+l$ values. Note that $k+l$ is the number of bit offsets within a page. Figure~\ref{fig:prob1} shows that, for 1 bit per page, 2200 pages are enough to achieve 99.99\% accuracy for the DDR4 DRAM K1 listed on Table~\ref{tab:flip_profiles}. For 2 and 3 bits per page, the same number of pages give 2\% and 0.006\% probability, respectively.
\begin{figure}[h]
    \centering
    \includegraphics[width=\columnwidth]{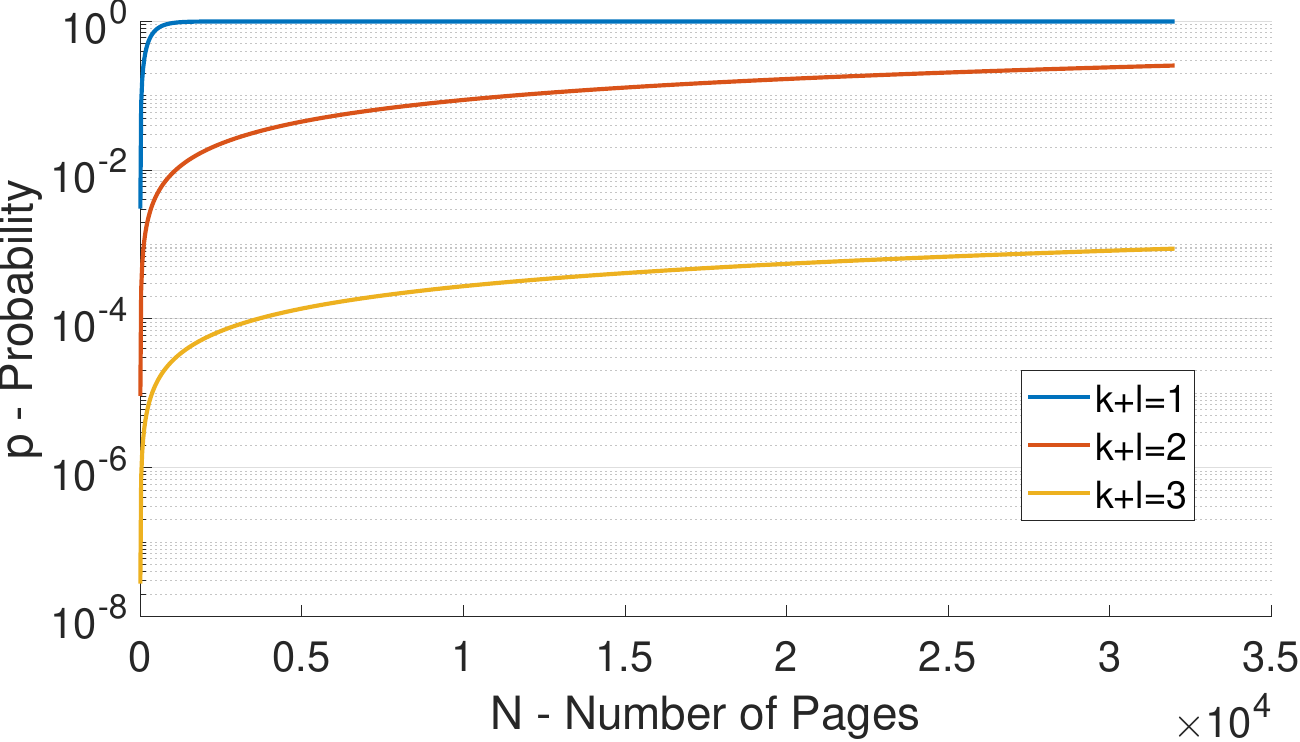}
    \caption{Probability of finding a page among N pages for different k+l values. k+l states the number of targeted bit offsets in a page.}
    \label{fig:prob1}
\end{figure}

Second, given a page offset to flip, we calculate the probability of finding such a target page for each $N$ values and different DRAM chips. The results on Figure~\ref{fig:prob2} suggest that given enough number of pages, 
$N$, the probability of finding a target page is close to 1 for even the least flippy DRAM devices.

\begin{figure}[h]
    \centering
    \includegraphics[width=0.96\columnwidth]{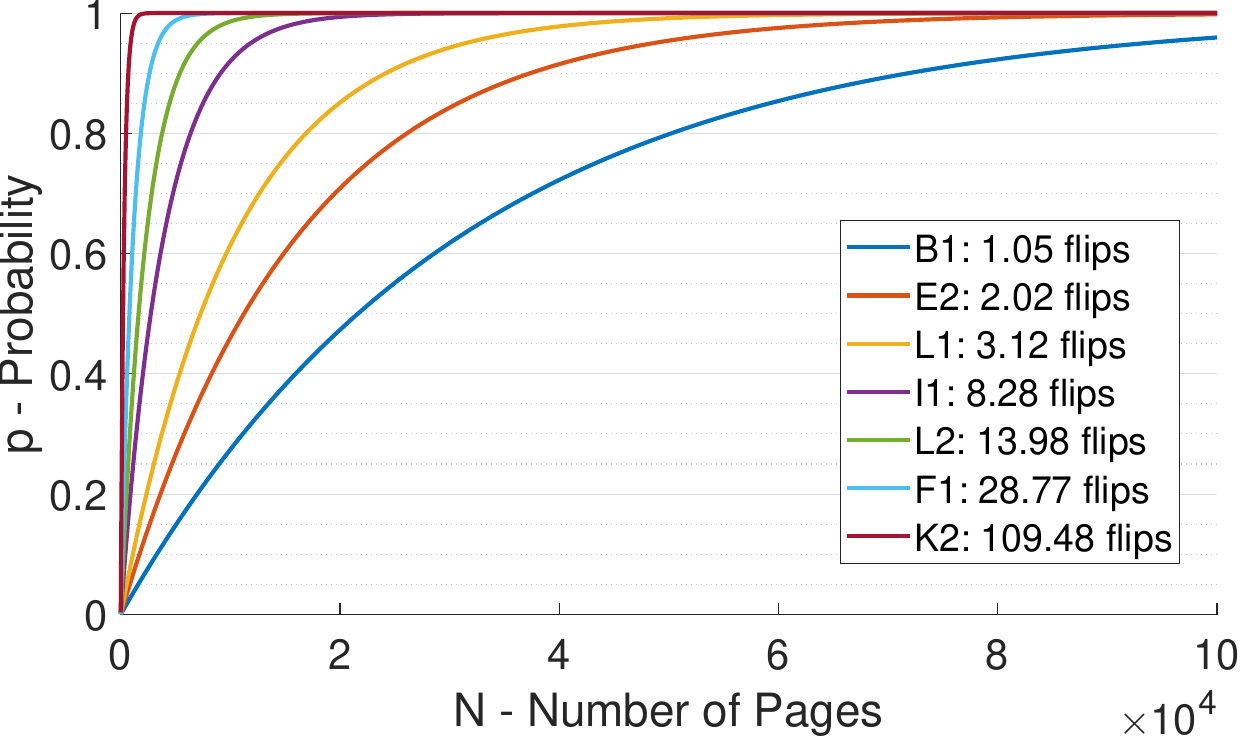}
    \caption{Probability of finding a page among N pages for different DRAM chips.}
    \label{fig:prob2}
\end{figure}

\subsection{Finding Contiguous Memory}\label{apx:spoiler}
Virtual to physical address mappings are stored in \textit{pagemap} file in Linux OSs and it requires root privileges to access these translations. Using the SPOILER tool~\cite{islam2019spoiler}, we can reliably leak the information about the last 8 bits in physical addresses after the page offset bits. This allows us to find contiguous memory chunks in physical address space. 

SPOILER works by taking advantage of a performance optimization in Intel processors where loads are executed speculatively before stores, resulting in a timing side channel if the load has dependencies based on the same partial address information as a store. SPOILER uses the differences in time between loads and stores to extrapolate which virtual addresses within an array are contiguous physically. Our SPOILER implementation performs the timing measurements 100 times per page, removing outliers and taking the average read time. 

In Figure~\ref{fig:spoiler_peaks}, the page numbers with the peaks in the y-axis are contiguous pages physical address space. 
\begin{figure}[h]
    \centering
    \includegraphics[width=\columnwidth]{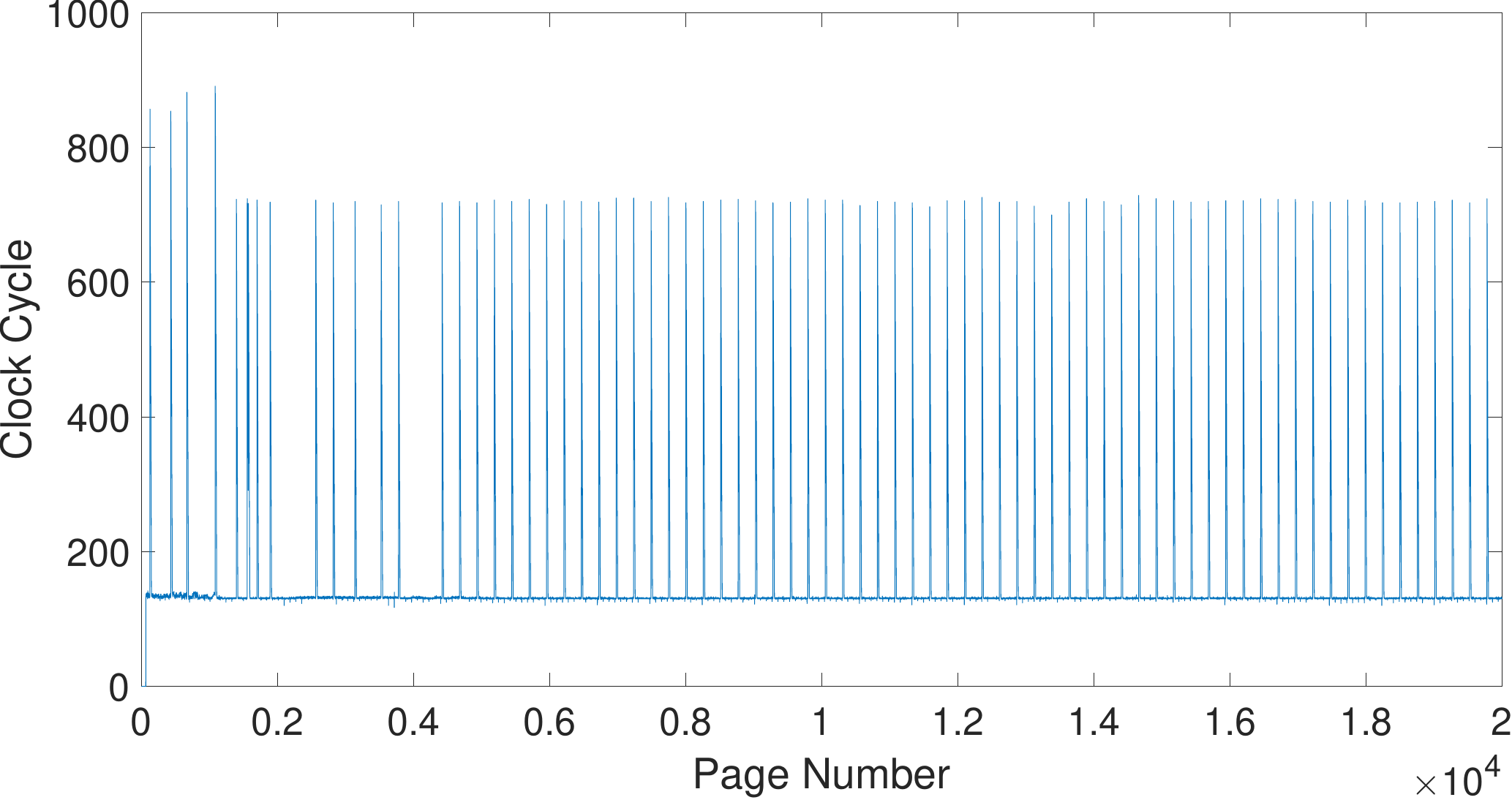}
    \caption{Timing peaks on virtual addresses detected by SPOILER~\cite{islam2019spoiler} attack. Virtual addresses on the peaks are contiguous on physical address space.}
    \label{fig:spoiler_peaks}
\end{figure}

\subsection{Finding Neighbor Rows in DRAM}\label{apx:row_conflict}
After translating the address from virtual to physical address space, there is another translation which maps the physical addresses to DRAM banks, rows and columns. To be able to successfully realize a Rowhammer attack, we need physical addresses that are mapped to the same bank in the DRAM, and thus, neighbor rows. We use row buffer conflict side-channel~\cite{pessl2016drama} to detect which two addresses are in the same bank. Figure~\ref{fig:row_conflict} shows that about one sixteenth of the address give larger access time, meaning they are in the same banks and neighbor rows.
\begin{figure}[h]
    \centering
    \includegraphics[width=\columnwidth]{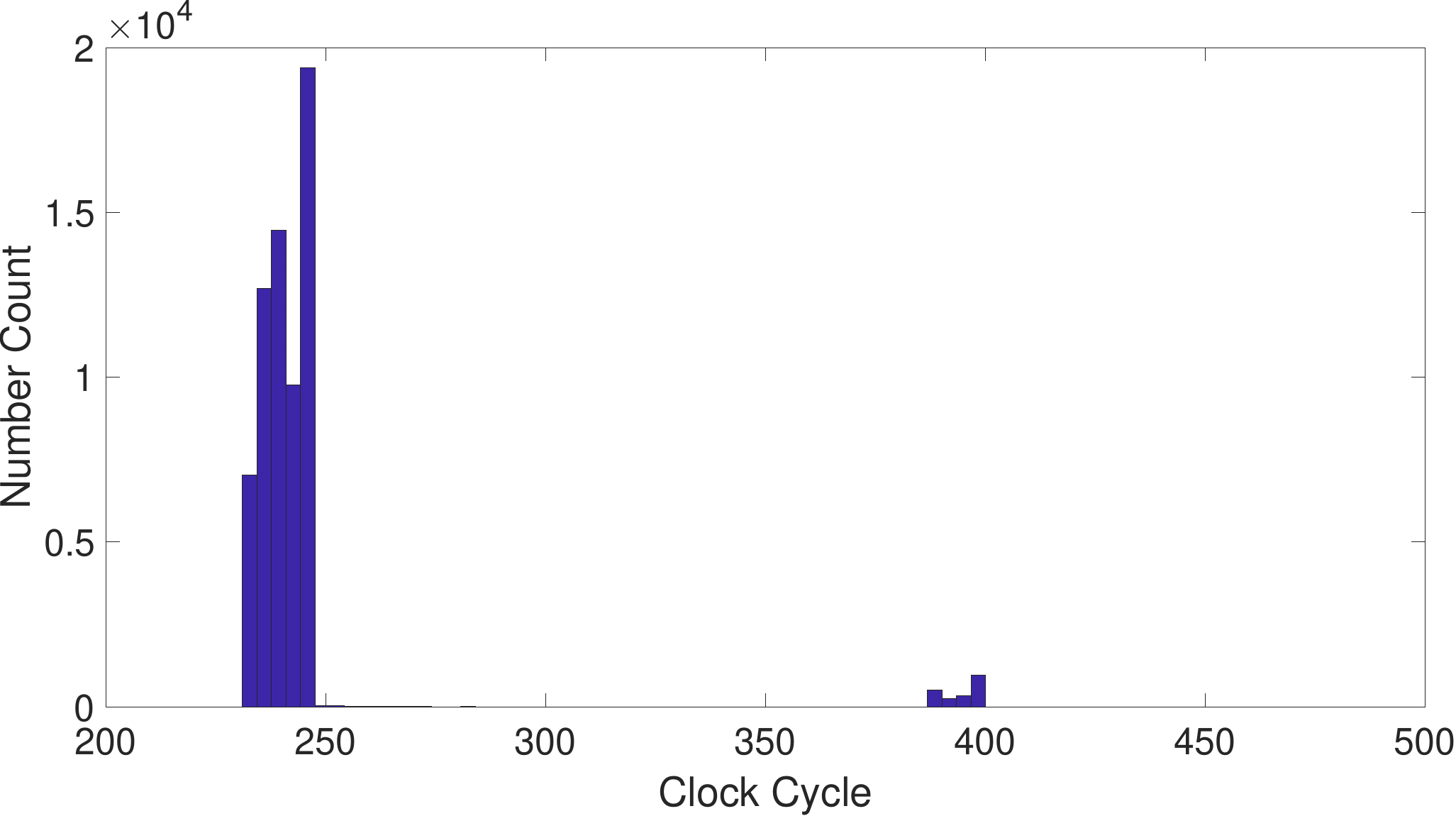}
    \caption{Access time distribution of previously found contiguous physical addresses. Accessing physical addresses that are mapped to the same DRAM bank takes around 400 clock cycles.}
    \label{fig:row_conflict}
\end{figure}

\subsection{Restoring the Modified Parameters}\label{sec:limiting}

In order to show the importance of putting constraints on the optimization of target neurons, first, we fine-tune all parameters in a ResNet18 model  using clean and adversarial examples without putting any constraints. Since the training approach is aligned with the work in~\cite{gu2017badnets}, we refer to this method as BadNet. Then, starting from the weight parameters with the lowest gradient values, we restore a part of the parameters to their original values at the end. 
Table~\ref{tab:ft_vs_cft} shows the change in the attack performance a part of the weight parameters are restored into their original values. For instance, BadNet reaches 99.88\% Attack Success Rate and 87.61\% Test Accuracy after the unconstrained fine-tuning. Then, when we restore only 1\% of the weights (i.e. 99\% of the weights remain modified), we observe that Attack Success Rate drops down to 76.11\% while the Test Accuracy is slightly increasing.
Even if we keep 50\% of the parameters (44 million bits) modified, limiting the number of modified parameters at the end of fine-tuning achieves only 34.15\% Attack Success Rate, whereas we reach $92.95\%$ test accuracy and $95.26\%$ attack success rate with only 99 bit-flips using CFT+BR. Therefore, we claim that fine-tuning without any constraints distributes the knowledge of backdoor to all parameters and the parameter limit that is applied at the end degrades the backdoor success rate drastically. This result pushes us towards putting constraints on fine-tuning.

\begin{table}[h]
    \centering
    \caption{BadNet reaches reasonable attack success rate (ASR) and test accuracy (TA) only when more than 90\% or parameters are changed. Limiting the percentage of modifications after fine-tuning decreases the attack performance.}
    \begin{tabular}{c|c|c}
    \toprule
         Modification(\%) & TA(\%) & ASR(\%)  \\
  \midrule
        100 & 87.61 & 99.88 \\
        99 & 89.79      &  76.11         \\
        90 & 90.92 &  61.04     \\
        80 & 91.67 & 51.22      \\
        70 & 92.01 & 43.79     \\
        50 & 92.41 & 34.15     \\
       \bottomrule
    \end{tabular}
    
    \label{tab:ft_vs_cft}
\end{table}

\subsection{Comparison of Found Bit Flips}\label{apx:flip_comparison}
Figure~\ref{fig:flip_comparison} illustrates the sparsity of found bit flip locations by our method (CFT+BR) and TBT. Note that the bit flips found by TBT (red) are localized in only one page. However, the bit flips located by CFT+BR are sparsely distributed over the weight file which makes them actually flippable in the online phase.

\begin{figure}[h]
\vspace{-0.53cm}
    \centering
    \includegraphics[width=\columnwidth]{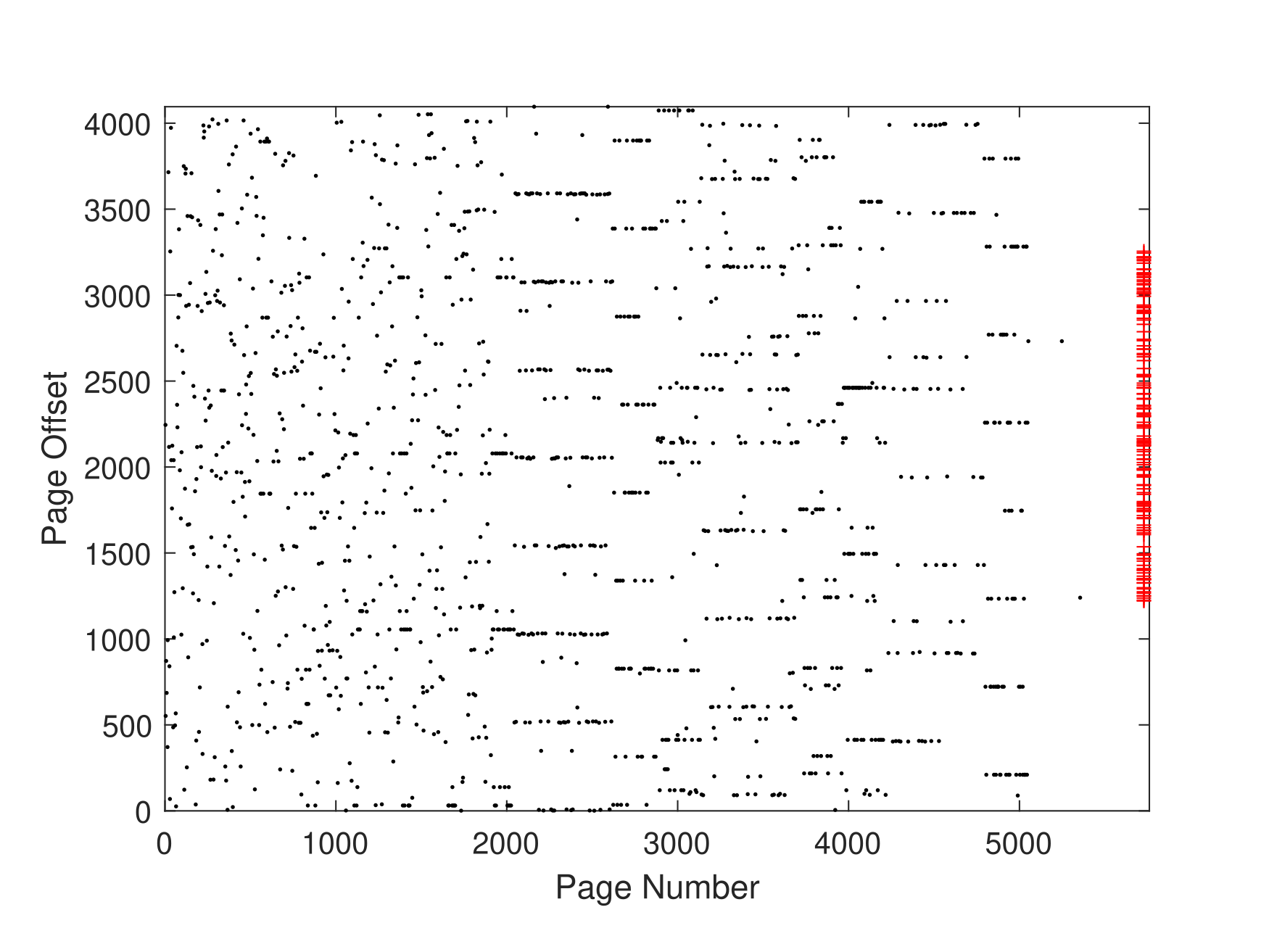}
    \vspace{-0.5cm}

    \caption{The comparison of vulnerable bit locations found by CFT+BR (black) and TBT (red) on ResNet50 weight file.}
    \label{fig:flip_comparison}
\end{figure}

\subsection{Negative Result - Plundervolt Attack}
In this experiment, we try to use another software-based faulting mechanism, namely Plundervolt~\cite{Murdock2019plundervolt}, to inject faults during the inference phase of a DNN model. Differently from the Rowhammer attack, the Plundervolt attack utilizes undervolting the CPU beyond the optimal operation limits using the MSR interface to cause faulty results in the multiplication results. Since the computational graphs of DNN models have many multiplication operations, Plundervolt can be a potential threat to DNN inference as well. We first run the Plundervolt PoC code to verify undervolting can fault the multiplication operations and get the frequency-voltage pair where we can reliably produce faults. Then, we experimented with DNN models with floating-point weight parameters and undervolted the CPU to the determined frequency-voltage pair. We did not observe any faults in the multiplication results when the operands are floating points. We also experimented undervolting while an n-bit quantized DNN model is operating. However, we did not see any faults in the DNN model. We claim that the reason why the multiplication results are not affected by undervolting is the operand values are limited to $2^n-1$ which is 255 in 8-bit quantized DNN models. We observed that when the second operand of multiplication is smaller than \texttt{0xFFFF}, undervolting does not introduce any bit flips in the multiplication result which is consistent with the observations in the original Plundervolt work~\cite{Murdock2019plundervolt}. 

We also experimented with the matrix multiplication implementations of the PyTorch library. We observe that \texttt{torch.matmul} function produces faulty results only when the following three conditions are met. First, the second operand must be larger than \texttt{0xFFFF}. Second, the size of operands must be 1-by-1. Finally, the multiplication operation must be run in a while loop keeping operands constant.

Hence, we conclude that injecting backdoors to the DNN models or degrading the accuracy of them using Plundervolt attack is not practical.

\end{document}